\documentclass[10pt,twocolumn,letterpaper]{article}

\usepackage[preprint]{cvpr}      

\newcommand{\src}{\mathcal{S}}
\newcommand{\tar}{\mathcal{T}}

\definecolor{cvprblue}{rgb}{0.21,0.49,0.74}
\usepackage[pagebackref,breaklinks,colorlinks,allcolors=cvprblue]{hyperref}
\usepackage[utf8]{inputenc} 
\usepackage[T1]{fontenc}    
\usepackage{hyperref}       
\usepackage{url}            
\usepackage{booktabs}       
\usepackage{amsfonts}       
\usepackage{nicefrac}       
\usepackage{microtype}      
\usepackage{xcolor}         
\usepackage{amsmath} 
\usepackage{multirow}
\usepackage{colortbl}
\usepackage{graphicx}
\usepackage{float}
\newcommand{\vfo}{DV-Matcher }

\newcommand{\pd}{\Pi_{\mathcal{D}}}
\newcommand{\sd}{\mathcal{S}_{\mathcal{D}}}

\def\paperID{2489} 
\def\confName{CVPR}
\def\confYear{2025}

\title{DV-Matcher: Deformation-based Non-Rigid Point Cloud Matching Guided by Pre-trained Visual Features } 
\author{
   Zhangquan Chen$^{1 \#}$ $\quad{}$  Puhua Jiang$^{1, 2 \#}$  $\quad{}$  Ruqi Huang$^{1}$\thanks{ $\#$ indicates equal contribution.  Corresponding author: ruqihuang@sz.tsinghua.edu.cn }\\
   1. Tsinghua Shenzhen International Graduate School, China $\quad{}$ 2. Pengcheng Laboratory, China 
}

\begin{document}

\maketitle
\begin{abstract}

In this paper, we present DV-Matcher, a novel learning-based framework for estimating dense correspondences between non-rigidly deformable point clouds. 
Learning directly from unstructured point clouds \emph{without meshing or manual labelling}, our framework delivers high-quality dense correspondences, which is of significant practical utility in point cloud processing. 
Our key contributions are two-fold: 
First, we propose a scheme to inject prior knowledge from pre-trained vision models into geometric feature learning, which effectively complements the local nature of geometric features with global and semantic information; 
Second, we propose a novel deformation-based module to promote the extrinsic alignment induced by the learned correspondences, which effectively enhances the feature learning. 
Experimental results show that our method achieves state-of-the-art results in matching non-rigid point clouds in both near-isometric and heterogeneous shape collection as well as more realistic partial and noisy data. The code for this project is available at \url{https://github.com/rqhuang88/DV-Matcher.git}.

\end{abstract}
\section{Introduction}\label{sec:intro}

Point cloud data has long been the most prevailing form of 3D data acquisitions (e.g., laser scanning, photogrammetry). 
In this paper, we propose \textbf{DV-Matcher}, a learning-based framework for matching \emph{non-rigidly deformable} point clouds -- a task that is essential for reconstructing~\cite{yu2018doublefusion}, understanding~\cite{scape} and manipulating~\cite{paravati2016point} \emph{dynamic} objects in real world.

\begin{figure}[t]
    \vspace{-1em}
    \centering
    \includegraphics[width=0.5\textwidth]{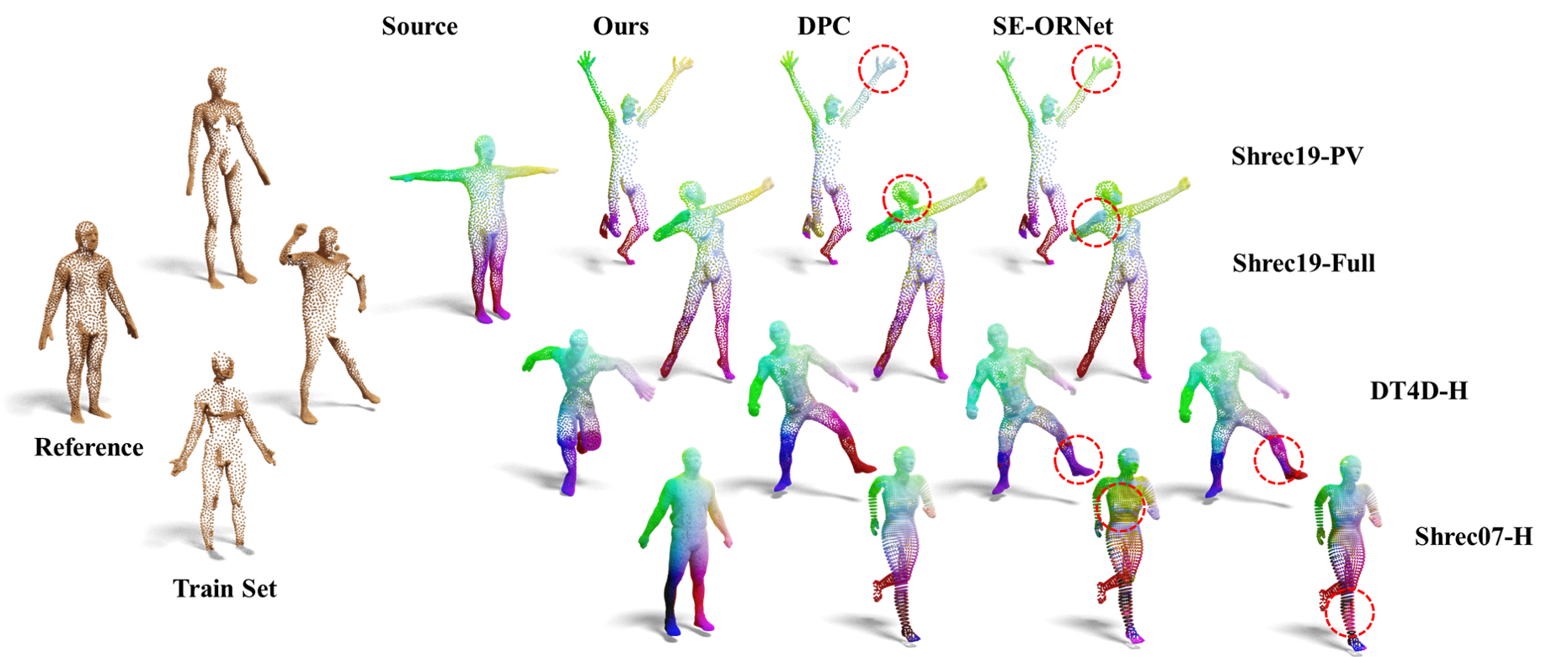}
    \vspace{-1em}
    \caption{
	We train DV-Matcher and two baselines under a challenging setting. The training set consists of \emph{only one full point cloud as reference and a set of $516$ partial point clouds} sampled from shapes in SHREC'19 with significant pose/style deformations (see the yellow point clouds). Without any correspondence label, our framework not only manages to match accurately on SHREC'19 benchmark (\emph{both} partial and full setting), but also generalizes well to unseen benchmarks. See Sec.~\ref{sec:realapp} for more details.
    }\label{fig:teaser}
    \vspace{-1em}
\end{figure}

Apart from the primary pursuit of accurate dense correspondences, we have extended our effort towards the following goals for boosting the utility and potential of our framework in real-world applications: 
G1) \textbf{Robustness} with respect to significant deformations and partiality; 
G2) \textbf{Efficiency} for scalability in processing large-scale data; 
G3) \textbf{Correspondence Label-Free} for minimizing manual effort from users; 
G4) \textbf{Preprocessing-Free} for direct data training/inference, without introducing extra operations such as meshing and point re-sampling. 
Though, as will be discussed soon, different combinations of the above goals have been achieved by the prior arts, \vfo demonstrates high capacity along \emph{all} axes, but also delivers state-of-the-art matching accuracy (see Fig.~\ref{fig:teaser} for a challenging example). 

In fact, estimating correspondences between non-rigid point clouds has long attracted research interests. 
Early approaches~\cite{nicp,liao2009modeling,li2008global}are mostly axiomatic and featured by lightweight and efficient implementation, which fulfill G2-G4 above. 
When the deformations among input are small to moderate, which is typical in dynamic 3D reconstruction~\cite{yu2018doublefusion, newcombe2011kinectfusion}, such algorithms are effective. 
On the other hand, due to their dependence on spatial proximities of input, they often struggle in the presence of large deformation in either pose/style~\cite{dfr}. 
Some recent advances aim for robustness regarding pose variation~\cite{zhao2024clustereg}, which on the other hand falls short of handling style variation and partiality (see Sec.~\ref{sec:exp}). 

To deal with large deformation, a recent trend~\cite{marin2020correspondence,nie,cao2023, dfr} is to transplant the success of matching triangular meshes via \emph{spectral method}~\cite{ovsjanikov2012functional} into the domain of point clouds. 
In essence, such approaches follow a self-supervised scheme built on the bijective mapping between a mesh and its vertex set, and therefore passing the \emph{intrinsic-geometry-aware} features to the point-based encoder. 
While achieving great performance, these methods all require triangular meshes during training. 
In fact, all the above methods are trained on existing mesh benchmarks, falling short of meeting G4.


In parallel, there exists another line of works~\cite{groueix20183d,lang2021dpc, zeng2021corrnet3d, deng2023se} learning to match without more advanced data representation. 
In particular, they leverage cross-reconstruction between point clouds induced by the estimated soft-maps as a proxy task for learning a feature extractor.  
While they satisfy G1-G4 to some extent, the reconstruction process does not physically align the input point clouds and suffers from mode collapsing, hindering achieving high-quality correspondences. 


The above observations indeed reveal two critical points in designing a strong correspondence estimator -- 1) \textbf{Easy-to-obtain matching cues other than spatial proximities} and 2) \textbf{A good proxy task for verifying and guiding correspondence learning}. 

For the former, instead of endowing surface geometry to point clouds (\emph{e.g., }meshing), we take an easier path of projecting them into 2D images. 
Then we aggregate the regarding semantic meaningful features extracted from multi-view projections by pre-trained vision models back to 3D, which empirically serve as high-quality matching cues. 
In fact, the idea of applying pre-trained vision models in shape analysis has recently attracted considerable attention~\cite{dutt2023diffusion, back3d, abdelreheem2023zero, snm}. 
Nevertheless, we emphasize that these methods essentially leverage pre-trained vision models to generate intermediate cues (\emph{e.g., }landmark correspondences) to assist 3D tasks, which can be error-prone. 
In contrast, our approach treats the aggregated features as visual encoding (similar to positional encoding~\cite{mildenhall2021nerf}), and trains a strong feature extractor guided by the following deformation-based proxy task. 

For the latter, we advocate correspondence-induced non-rigid shape deformation as the proxy task of choice. 
While the advantage of such proxy task is obvious, the prior arts~\cite{dfr,feng2021recurrent} based on deformation take an iterative optimization approach, which prevents efficient inference (G2). 
On the other hand, the existing end-to-end deformation frameworks~\cite{trappolini2021shape, huang20} require correspondence supervision (G3).  
In contrast, thanks to the visual encoding as well as tailored-for network design for exploit both visual and geometric information (see Sec.~\ref{sec:method}), our DV-Matcher achieve both accuracy and efficiency in deformation-based feature learning without correspondence label.


We conduct a rich set of experiments to verify the effectiveness of our pipeline, highlighting that it achieves state-of-the-art results in matching non-rigid point clouds under various settings including near-isometric, heterogeneous, full, partial and even realistic scanned point clouds from a range of categories. 
Remarkably, it generalizes well despite the distinctiveness between the training set and test set. 

\section{Related Works}
\subsection{Non-rigid Shape Matching}
Non-rigid shape matching is a long-standing problem in computer vision and graphics.  Unlike the rigid counterpart, non-rigidly aligning shapes is more
challenging owing to the complexity inherent in deformation models. 

Originating from the foundational work on functional maps~\cite{ovsjanikov2012functional}, along with a series of follow-ups~\cite{nogneng2017informative, huang2017adjoint, BCICP, zoomout, huang2020consistent,litany2017deep,diffusionNet,cao2022, li2022attentivefmaps, donati2022deep,attaiki2021dpfm,dual}, spectral methods have made significant progress in addressing the non-rigid shape matching problem, yielding state-of-the-art performance. However, because of the heavy dependence of Laplace-Beltrami operators, DFM can suffer notable performance drop when applied to point clouds without adaptation~\cite{cao2023}. In fact, inspired by the success of DFM, several approaches~\cite{nie, cao2023,dfr} have been proposed to leverage intrinsic geometry information carried by meshes in the training of feature extractors tailored for non-structural point clouds.  When it comes to pure point cloud matching, there is a line of works~\cite{zeng2021corrnet3d,lang2021dpc,deng2023se} leverage point cloud reconstruction as the proxy task to learn embeddings without correspondence labels. Since intrinsic information is not explicitly formulated in these methods, they can suffer from significant intrinsic deformations and often generalize poorly to unseen shapes.

\subsection{Pre-trained Vision Model for Shape Analysis}
 Recently, Pre-trained Vision Models have become increasingly popular in due to their remarkable ability to understand data distributions from extensive image datasets. In the fields of shape analysis, ~\cite{i2p} proposes an alternative to obtain superior 3D representations from 2D pre-trained models via Image-to-Point Masked Auto-encoders. ~\cite{abdelreheem2023zero} introduces a fully multi-stage method that exploits the exceptional reasoning capabilities of recent foundation models in language~\cite{openai2021gpt3} and vision\cite{li2023blip} to tackle difficult shape correspondence problems. In ~\cite{snm}, before surface matching, the authors propose to use the features extracted from DINOv2~\cite{oquab2023dinov2} of multi-view images of the shapes to perform co-alignment. In contrast to these approaches, which primarily utilize coarse patch features for sparse landmarks or semantic matching, our approach introduces an end-to-end method that aggregates pixel-level 2D features into point-wise 3D features.

\subsection{Non-rigid Partial Shape Matching}
While significant advancements have been made in full shape matching, there remains considerable room for improvement in estimating dense correspondences between shapes with partiality. Functional maps representation~\cite{rodola2016partial,attaiki2021dpfm,cao2023} has already been applied to partial shapes. However, both axiomatic and learning-based lines of work typically assume the input to be a \emph{connected mesh}, with the exception of~\cite{cao2023}, which relies on graph Laplacian construction~\cite{sharp2020laplacian} in its preprocessing. For partial point cloud matching, axiomatic registration approaches ~\cite{nicp,amm,li2022non} assume the deformation of interest can be approximated by local, small-to-moderate, rigid deformations, therefore suffer from large intrinsic deformations. Simultaneously, there's a growing trend towards integrating deep learning techniques~\cite{bozic2020deepdeform, bozic2020neural, huang_multiway_2022,li2022lepard}. However, these methods often focus on addressing the partial sequence point cloud registration problem.

\section{Methodology}\label{sec:method}

\begin{figure}[t!]
    \centering
    \includegraphics[width=0.5\textwidth]{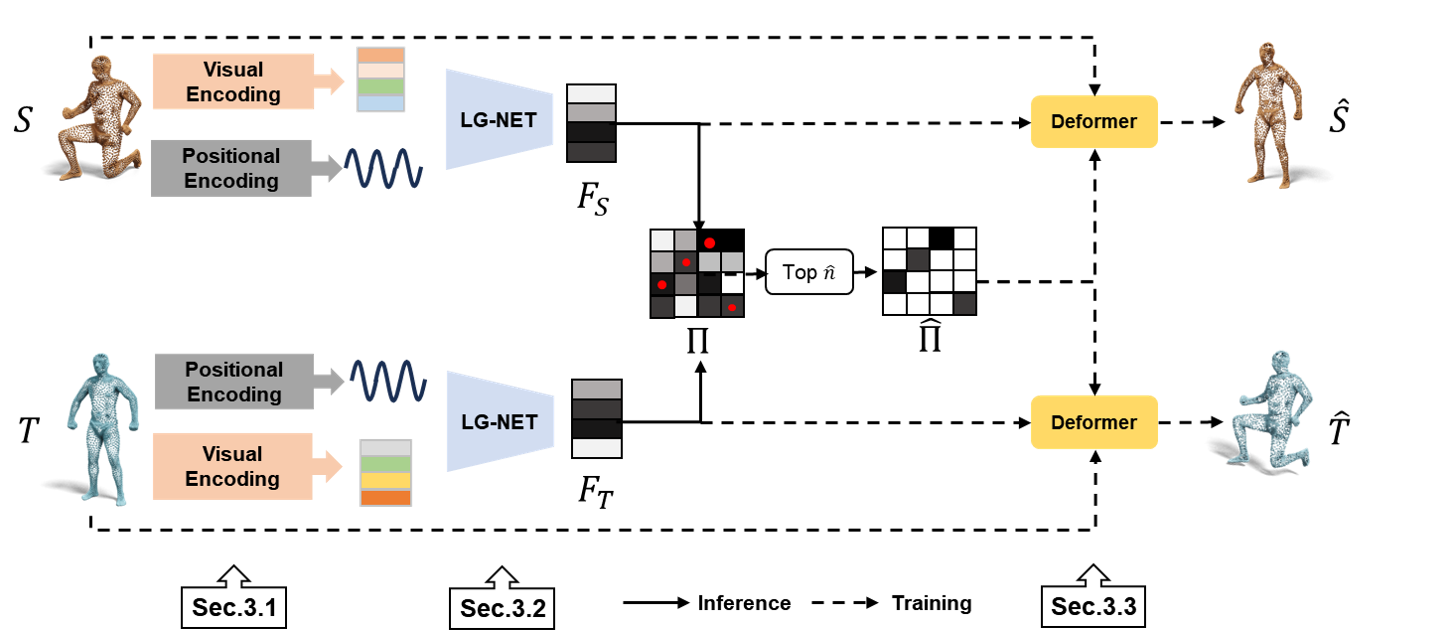}
    \vspace{-1.5em}
    \caption{The schematic illustration of our pipeline. 
    }\label{fig:pipeline}
    \vspace{-1.5em}
\end{figure}

Fig.~\ref{fig:pipeline} shows the overall pipeline. 
We first introduce our visual encoding method in Sec.\ref{sec:per-point}. 
Then our global and local attention network will be discussed in Sec.\ref{sec:network}. 
The training losses are described in Sec.\ref{sec:loss}.

\subsection{Visual Encoding}\label{sec:per-point}
Given a point cloud $P$ consisting of $N$ points, we denote the $i-$th point by $p_i = (x_i, y_i, z_i)$, our goal is to obtain a per-point feature carrying semantic information from pre-trained visual models. For more details, please refer to the Supp. Mat.

\noindent\textbf{Depth aware projection}: 
Following I2P-MAE~\cite{i2p}, we project $P$ on $xy-, yz-, xz-$plane to obtain three images. F
Taking $xy-$plane for example, we project point $p_i$ onto pixel $(u_i, v_i) = (\lfloor \frac{x_i - x_{\min}}{\Delta}\times H\rfloor, \lfloor \frac{y_i - y_{\min}}{\Delta}\times W\rfloor)$, with pixel intensity $f(u_i, v_i) = sigmod(z_i)$. 
Here $\Delta = \max\{x_{\max}-x_{\min}, y_{\max} - y_{\min}\}$, and $H, W$ are pre-determined image dimensions. 
For unprojected pixels, we simply set intensity to $0$.


We note that the projected images often come with holes due to the discrete nature of point clouds, which are distinctive from the realistic training images used in DINO. 
To alleviate the discrepancy, we propose to 1) apply a $3\times 3$ mean filter on the gray images and 2) assign pseudo color on the pixel values with the \texttt{PiYG} colormap in MATLAB. 
We now denote by $I_{\hat{z}}, I_{\hat{x}}, I_{\hat{y}}$ to resulting images, where $\hat{z}$ indicates projection onto $xy-$ plane (and similarly for $\hat{x}, \hat{y}$). 


\noindent\textbf{Lifting image features}:
Directly applying DINOv2 on the projected images results in coarse features of dimension $D\times D \times C$, where $D$ is small (\emph{e.g.,} 16). 
We further leverage FeatUp~\cite{fu2024featup} to upsample DINOv2 features to match the dimension of the projected images
\begin{equation}
    F^{img}_{\hat{z}} = \Theta(I_{\hat{z}})\in \mathbb{R}^{H\times W\times C},     
\end{equation}
where $\Theta$ is the per-pixel encoder of DinoV2-FeatUp~\cite{fu2024featup} and $C$ is the number of channels for each pixel. 
Via the one-to-one correspondences between $p_i$ and $(u_i, v_i)$, we obtain the point-wise feature of $p_i$ via a simple pull-back:
\begin{equation}
    f^i_{\hat{z}} = F_{\hat{z}}(u_i, v_i, :)\in \mathbb{R}^{C}. 
\end{equation}
We then have $F^{pt}_{\hat{z}}\in \mathbb{R}^{N\times C}$ by stacking $f^i_{\hat{z}}$ in order. 
We compute $F^{pt}_{\hat{x}}, F^{pt}_{\hat{y}}$ in the same manner. We emphasize that these computations are independent. In the end, we arrive at 
\begin{equation}\label{eq:pfeat}
F^{pt}(P) = [F^{pt}_{\hat{z}}, F^{pt}_{\hat{x}}, F^{pt}_{\hat{z}}] \in \mathbb{R}^{N\times 3C}.    
\end{equation}
The above procedure returns a set of per-point features for the input $P$, which essentially carry the semantic information extracted by the visual encoding.  

\subsection{Local and Global Attention Network}\label{sec:network}

In this part, we describe our Local and Global Attention Network, which are depicted in Supp. Mat.

\noindent\textbf{Input feature: }
In order to exploit both visual (image-based) and geometric (point-based) features, we perform early fusion at the input stage as follows: 
\begin{equation}\label{eq:ifeat}
    F^{in}(P) = \mbox{LBR}(F^{pt}(P)) + \gamma(P),
\end{equation}
where $\gamma(P)\in \mathbb{R}^{N\times 384}$ is the positional encoding~\cite{mildenhall2021nerf} and LBR is a module proposed in PCT~\cite{guo2021pct} for non-linearly converting $F^{pt}(P)$ into the same dimension of $\gamma(P)$. 



\noindent\textbf{Architecture design:} 
We propose a dual-pathway architecture in parallel, comprising global attention~\cite{guo2021pct} and local attention~\cite{wu2023attention} blocks, which effectively draw matching cues from different hierarchical levels. 

However, the receptive filed of local attention in~\cite{wu2023attention} is pre-computed using point coordinates and fixed through training, which can be misleading in non-rigid shape matching. 
For instance, one's hand can be close to his head, but their intrinsic distance should always be large.  
Inspired by DGCNN~\cite{dgcnn}, we lift the neighborhood search the feature domain and keep updating it through learning.  

In the end, we propose a fusion module consisting of LBR and a three-layer stacked N2P~\cite{wu2023attention} attention, to merge features from both global and local paths, resulting in our output feature. 
We refer readers to the Supp. Mat. for more details.

\subsection{Training Objectives and Matching Inference}\label{sec:loss}

In the following, we introduce our training losses, which consist of our novel deformation-based loss, arap loss, smoothness loss and geometrical similarity loss. 
As shown in Fig.~\ref{fig:pipeline}, our main model is a Siamese network. 
Given a pair of point clouds $\src\in \mathbb{R}^{N\times 3}, \tar \in  \mathbb{R}^{M\times 3}$, we compute $C$-dimension per-vertex features $F_{\src} \in \mathbb{R}^{N\times C}, F_{\tar}\in  \mathbb{R}^{M\times C}$ 
from the LG-Net (Sec.~\ref{sec:network}). 
We then estimate dense correspondences, $\Pi_{\src\tar} \in \mathbb{R}^{N \times M}$ and $\Pi_{\tar\src} \in \mathbb{R}^{N \times M}$, with respect to Euclidean distance~\cite{cover1967nearest} among rows of $F_{\src}$ and $F_{\tar}$, followed by a softmax normalization. 
Then, we select the matches set with top $\hat{n} = 10$ matching scores and set the remaining elements to zero to get $\hat{\Pi}_{\src\tar}$ and $\hat{\Pi}_{\tar\src}$, which are used to define the following losses. 

\begin{figure}[!t]
    \centering
    \includegraphics[width=0.5\textwidth]{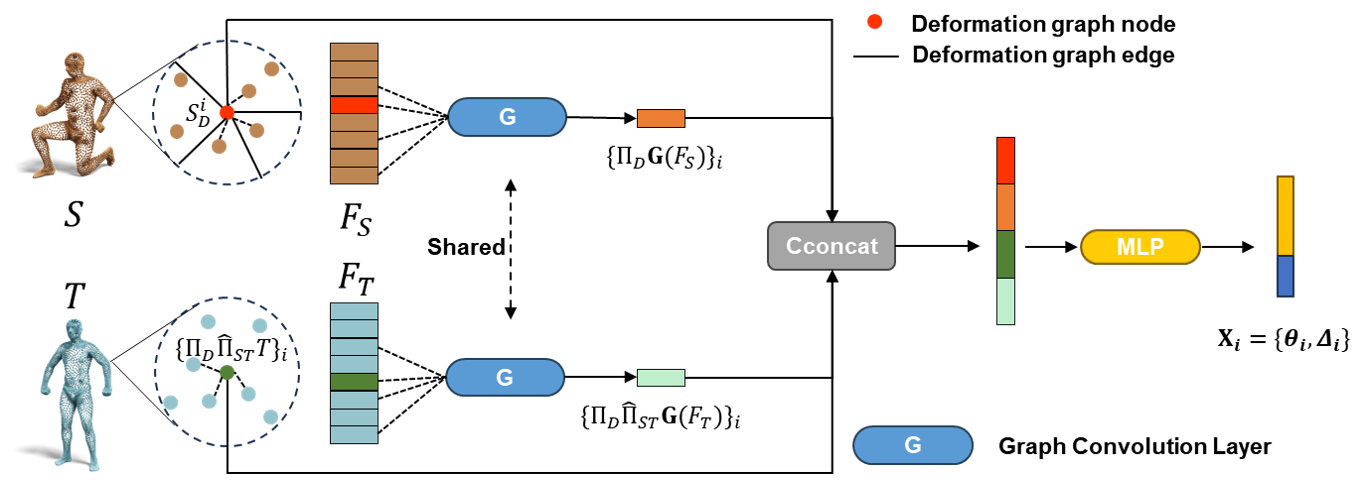}
    \caption{Illustration of our deformer, which predicts rigid transformation at each deformation graph node.}\label{fig:deformer}
    \vspace{-1.5em}
\end{figure}

\noindent\textbf{Deformation-based loss} aims to deform one point cloud to the other using the above predicted $\hat{\Pi}_{\src\tar}$ and $\hat{\Pi}_{\tar\src}$ as shown in Fig.~\ref{fig:deformer}. 

Taking the direction from $\src$ to $\tar$ for example, we start by constructing deformation graph on $\src$. 
Following~\cite{guo2021human,dfr}, we perform farthest point sampling (FPS) on $\src$ to obtain $[N/2]$ points, $\sd \in \mathbb{R}^{[N/2]\times 3}$, as nodes of the deformation graph. 
In particular, we encode the above sampling process as a binary matrix $\pd \in \mathbb{R}^{[N/2]\times N}$ such that $\pd(i, j) = 1$ if and only if the $j$-th point of $\src$ is sampled in the $i-$th round of FPS, and $\pd(i, j) = 0$ otherwise.
It is therefore evident that $\sd = \pd \src$. 

We then assign with each node in $\sd$ a rigid transformation parametrized as $\{\theta, \delta\}$, where $\theta$ is a 6-dim representation of rotation matrix~\cite{zhou2019continuity} and $\delta$ is a 3-dim translation vector. 
Stacking all together, we arrive at $\textbf{X} = \{\Theta, \Delta\}$, where $\Theta\in \mathbb{R}^{[N/2] \times 6}, \Delta \in \mathbb{R}^{[N/2] \times 3}$. 
Given $\textbf{X}$, we can then propagate the transformations from nodes in $\sd$ to each point in $\src$ via a distance-based weighting scheme, we refer readers to the Supp. Mat. for more details. 
Finally, we denote by $\hat{\src}$ the deformed version of $\src$ with respect to $\mathbf{X}$.
\begin{equation}\label{eq:dg}
\hat{\src}=\mathcal{DG}\left(\mathbf{X}, \src\right).
\end{equation}

In prior works~\cite{nicp, amm, dfr}, $\textbf{X}$ is often optimized jointly with correspondences in alternative iterations. 
In contrast, we propose to train a neural network to predict $\textbf{X}$ in a single feedforward.  
Specifically, we first construct a dimension-preserving graph convolution network $\mathbf{G}:\mathbb{R}^{\cdot \times C} \rightarrow \mathbb{R}^{\cdot \times C}$ for gathering information from $F_\src, F_\tar$, namely, the features learned from LG-Net. 
Then we use the following MLP to predict $\mathbf{X}$:
\begin{equation}\label{eq:mlp}
\mathbf{X}=\textbf{MLP}(\sd, \pd \mathbf{G}(F_\src), 
\pd \hat{\Pi}_{\src\tar} \tar, \pd \hat{\Pi}_{\src\tar} \mathbf{G}(F_\tar)).
\end{equation}
In the other words, we first pull back the point cloud $\tar$ and regarding feature $\mathbf{G}(F_\tar)$ to the source shape via $\hat{\Pi}_{\src\tar}$, then we perform the sampling regarding deformation nodes, $\pd$, on both spatial positions and latent features regarding both source and (pull-back) target shape. 
Finally, we predict $\mathbf{X}$ with all the above information. 

Now we define the deformation loss regarding $\mathbf{X}$ as the chamfer distance between deformed $\src$ and $\tar$: 
\begin{equation}
\mathcal{L}_{\mbox{deform}}^{(\src,\tar)}(\mathbf{X}) = CD(\mathcal{DG}(\mathbf{X}, \src), \tar),
\label{eq:cd}
\end{equation}
where $\mathbf{X}$ is learned via Eqn.~\ref{eq:mlp}. 

\noindent\textbf{ARAP loss: }
We adopt the classic As-Rigid-As-Possible (ARAP) regularization~\cite{guo2021human, levi2014smooth} on $\mathbf{X}$. 
We postpone the exact form of $\mathcal{L}^{(\src,\tar)}_{\mbox{arap}}$ to the Supp. Mat.

\noindent\textbf{Smoothness loss:} On the other hand, in order to encourage smoothness on the learned correspondences, we follow~\cite{lang2021dpc} to pose the following smoothness regularization: 
\begin{equation}
	\mathcal{L}^{(\src,\tar)}_{\mbox{smooth}} = CD(\tar,\hat{\Pi}_{\src\tar}\tar)
\end{equation}

\noindent\textbf{Geometrical similarity:} 
Since our predicted deformation graph comes from feature embedding $F_\src, F_\tar$,  we hope that the latter respects the local intrinsic geometry of the underlying surfaces. 
One high-level intuition is to enforce the Euclidean metric derived from the learned embeddings to approximate the underlying surface metric. 
Though this idea has been exploited in NIE~\cite{nie}, instead of minimizing the \emph{global absolute residual} between the two metrics, we opt for maximizing the \emph{local angular similarity} as follows. 

We first adopt the heat method~\cite{Crane:2017:HMD} to compute approximated geodesic matrix $\mathbf{M}_{\mathcal{S}}$ on $\src$.  
Then, treating $F_\src \in \mathbb{R}^{N\times C}$ as a $C-$dim embedding of $\src$, for each $x_i\in \src$, we compute its nearest neighbors in the embedded space and obtain $\mbox{NN}(i) = \{j_1, j_2, \cdots, j_k\}$ be the set of ordered indices. 
The regarding ascending list of distances is denoted by $d^i_{\src} \in \mathbb{R}^{k}$. 
On the other hand, we retrieve approximated geodesic distances from $\mathbf{M}_{\mathcal{S}}$ as $m_{\src}^i\in \mathbb{R}^k$, where $m_{\src}^i(t) = \mathbf{M}_{\mathcal{S}}(i, j_t), t = 1, 2, \cdots, k.$. The geometrical similarity loss can be denoted as:
\begin{equation}\label{eq:gs_loss}
    \mathcal{L}^{(\src)}_{\mbox{geo}} = \frac{1}{N}\sum_{i= 1}^N(1 - \frac{d_{\src}^i\cdot m_{\src}^i}{\Vert d_{\src}^i\Vert \Vert m_{\src}^i\Vert}).
\end{equation}
It is worth noting that the overhead of computing $\mathbf{M}_{\mathcal{S}}$ is only needed during training.




To summarize, we define the loss regarding the direction from $\src$ to $\tar$ as 
\begin{equation}\label{eq:loss}
\begin{aligned}
	\mathcal{L}^{(\src,\tar)}_{\mbox{total}} =& \lambda_{\mbox{deform}}\mathcal{L}^{(\src,\tar)}_{\mbox{deform}} + \lambda_{\mbox{arap}}\mathcal{L}^{(\src,\tar)}_{\mbox{arap}} \\
	&+ \lambda_{\mbox{smooth}}\mathcal{L}^{(\src,\tar)}_{\mbox{smooth}} +\lambda_{\mbox{geo}}\mathcal{L}^{(\src)}_{\mbox{geo}}. 
\end{aligned}
\end{equation}
Note that we perform training in a pairwise manner, we formulate all the above loss terms in both directions, that is
\[\mathcal{L}_{\mbox{total}} = \mathcal{L}^{(\src,\tar)}_{\mbox{total}} + \mathcal{L}^{(\tar,\src)}_{\mbox{total}}.\] 

\noindent\textbf{Partial matching loss:}
In the above, we entail the losses for training full-to-full non-rigid point cloud matching. 
Remarkably, our formulation can be easily extended to the challenging scenario of partial-to-full matching by modifying $\mathcal{L}_{\mbox{deform}}$ to a unilateral loss, i.e. only the partial-to-full chamfer distance is considered. 

\noindent\textbf{Inference: }
At inference time, we choose the nearest latent cross-neighborhood of $x_i \in \mathcal{S}$ to be its corresponding point by KNN \cite{cover1967nearest}, thus get the shape matching result between point cloud $\mathcal{S}$ and $\mathcal{T}$.

\subsection{Remarks}
\begin{figure}[!t]
    \centering
    \includegraphics[width=0.5\textwidth]{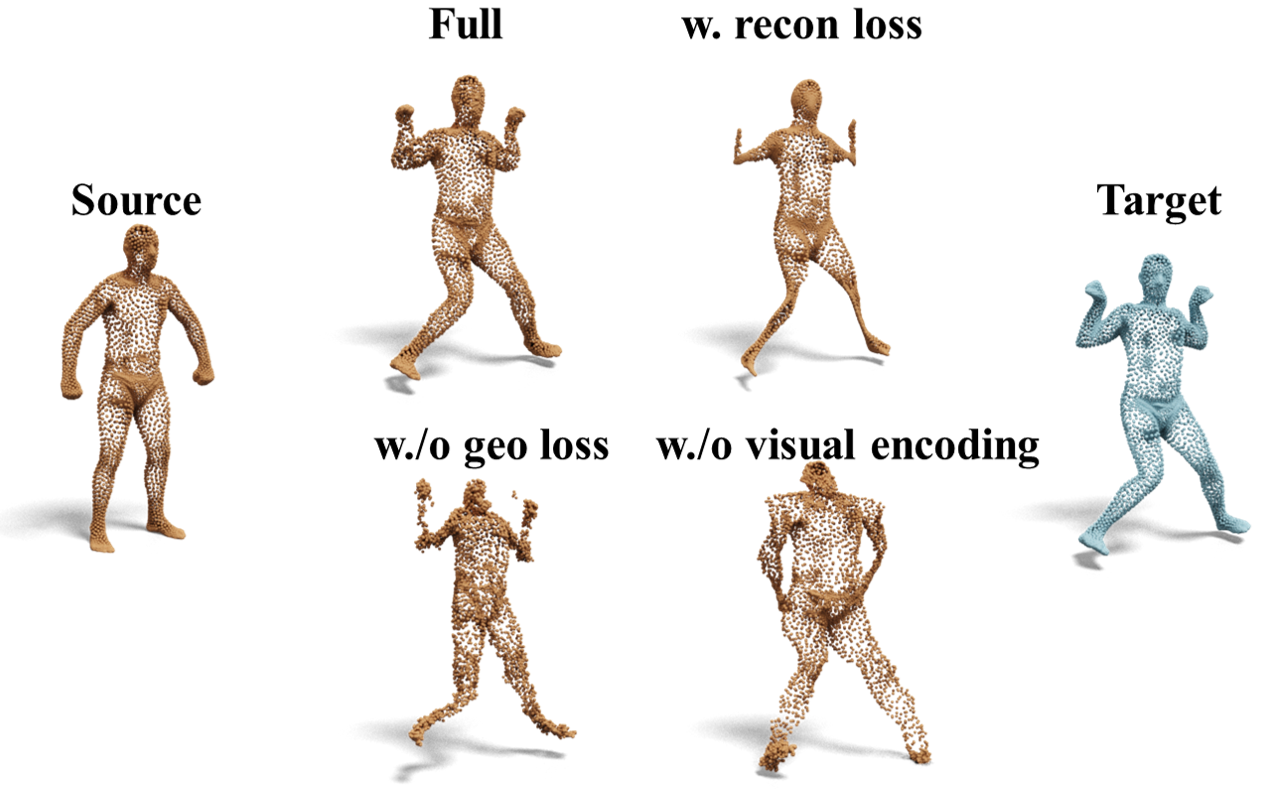}
    \caption{Ablation study on our training losses.}\label{fig:abl}
    \vspace{-1.5em}

\end{figure}
\begin{enumerate}
	\item While our exploitation of pre-trained vision models bears certain similarity to I2P-MAE~\cite{i2p}, we highlight the key difference -- I2P-MAE enforces point-based model to align with the vision model, as the features of vision model are also used as reconstruction target; on the other hand, our framework only use vision model for pre-encoding, the point model is essentially guided by geometry-based losses. 
	\item As mentioned in Sec.~\ref{sec:intro}, the reconstruction-based proxy task used in~\cite{lang2021dpc,deng2023se} can lead to mode collapse. We verify this in Fig.~\ref{fig:abl}, where "w. recon loss" means incorporating the above proxy task in our framework. Evidently, our deformation-based approach is free of collapsing. We also qualitatively justify our other designs in Fig.~\ref{fig:abl}.
\end{enumerate}

\section{Experiments}\label{sec:exp}

\begin{table*}[]
\caption{Quantitative results on SCAPE\_r (S\_r), FAUST\_r (F\_r), SHREC'19\_r (S19\_r), DT4D-H and SHREC07-H (S07-H) in terms of mean geodesic errors $(\times100)$. The \textbf{best} results from the pure point cloud methods in each column are highlighted.}\label{table:iso_noiso}
\vspace{-0.8em}
\centering
\footnotesize
\begin{tabular}{lcccccclccccc}
\hline
\multicolumn{1}{c}{}                                  & \textbf{Train}                                                              & \multicolumn{5}{c}{\textbf{S\_r}}                                                               &  & \multicolumn{5}{c}{\textbf{F\_r}}                                                               \\ \cline{3-7} \cline{9-13} 
\multicolumn{1}{c}{\multirow{-2}{*}{\textbf{Method}}} & \textbf{Test}                                                               & \textbf{S\_r} & \textbf{F\_r} & \textbf{S19\_r} & \textbf{DT4D-H}  & \textbf{S07-H} &  & \textbf{F\_r} & \textbf{S\_r} & \textbf{S19\_r} & \textbf{DT4D-H}  & \textbf{S07-H} \\ \hline
3D-CODED{[}S{]}~\cite{groueix20183d}                                      &                                                                             & 31.0              & 33.0              & \textbackslash{}    & \textbackslash{} & \textbackslash{}   &  & 2.5               & 31.0              & \textbackslash{}    & \textbackslash{} & \textbackslash{}   \\
TransMatch{[}S{]}~\cite{trappolini2021shape}                                    &                                                                             & 18.6              & 18.3              & 38.8                & 25.3             & 31.2               &  & 2.7               & 33.6              & 21.0                & 26.7             & 25.3               \\
NIE{[}U{]}~\cite{nie}                                           &                                                                             & 11.0              & 8.7               & 15.6                & 12.1             & 13.4               &  & 5.5               & 15.0              & 15.1                & 13.3             & 15.3               \\
SSMSM{[}U{]}~\cite{cao2023}                                         &      \multirow{-4}{*}{\begin{tabular}[c]{@{}c@{}}Mesh Required\end{tabular}}                                                                       & 4.1               & 8.5               & 7.3                 & 8.0              & 37.7               &  & 2.4               & 11.0              & 9.0                 & 11.8             & 42.2               \\ \hline
NDP{[}A{]}~\cite{li2022non}                                                    &                                                                             & 16.2              & \textbackslash{}  & \textbackslash{}    & \textbackslash{} & \textbackslash{}   &  & 20.4              & \textbackslash{}  & \textbackslash{}    & \textbackslash{} & \textbackslash{}   \\
AMM{[}A{]}~\cite{amm}                                                    &                                                                             & 13.1              & \textbackslash{}  & \textbackslash{}    & \textbackslash{} & \textbackslash{}   &  & 14.2              & \textbackslash{}  & \textbackslash{}    & \textbackslash{} & \textbackslash{}   \\
PointSetReg{[}A{]}~\cite{zhao2024clustereg}                                            &                                              & 17.1              & \textbackslash{}  & \textbackslash{}    & \textbackslash{} & \textbackslash{}   &  & 18.3              & \textbackslash{}  & \textbackslash{}    & \textbackslash{} & \textbackslash{}   \\
DiffFMaps{[}S{]}~\cite{marin2020correspondence}                                     &                                                                             & 12.0              & 12.0              & 17.6                & 15.9             & 15.4               &  & 3.6               & 19.0              & 16.4                & 18.5             & 16.8               \\
SyNoRiM{[}S{]}~\cite{multi}                                       &                                                                             & 9.5               & 24.6              & \textbackslash{}    & \textbackslash{} & \textbackslash{}   &  & 7.9               & 21.9              & \textbackslash{}    & \textbackslash{} & \textbackslash{}   \\
CorrNet3D{[}U{]}~\cite{zeng2021corrnet3d}                                     &                                                                             & 58.0              & 63.0              & \textbackslash{}    & \textbackslash{} & \textbackslash{}   &  & 63.0              & 58.0              & \textbackslash{}    & \textbackslash{} & \textbackslash{}   \\
DPC{[}U{]}~\cite{lang2021dpc}                                           &                                                                             & 17.3              & 11.2              & 28.7                & 21.7             & 17.1               &  & 11.1              & 17.5              & 31.0                & 13.8             & 18.1               \\
SE-ORNet{[}U{]}~\cite{deng2023se}                                      & \multirow{-6}{*}{PCD Only}                                                  & 24.6              & 22.8              & 23.6                & 27.7             & 12.2               &  & 20.3              & 18.9              & 23.0                & 12.2             & 20.9               \\
\rowcolor[HTML]{E7E6E6} 
Ours{[}U{]}                                          & \textbf{}                                                                   & \textbf{6.2}      & \textbf{5.1}      & \textbf{7.2}        & \textbf{6.9}     & \textbf{7.7}       &  & \textbf{5.4}      & \textbf{10.4}     & \textbf{9.3}        & \textbf{8.1}     & \textbf{8.2}       \\ \hline
\end{tabular}
\vspace{-1.5em}
\end{table*}

\noindent\textbf{Dataset:} We evaluate our method with several state-of-the-art methods on an array of different categories as follows, we defer the detailed descriptions of each benchmark to the Supp. Mat. for completeness.  
\begin{enumerate}
	\item Human: We consider the well-known benchmarks, including near-isometric ones -- \textbf{SCAPE\_r, FAUST\_r, SHREC'19\_r}, and more heterogenous ones -- \textbf{DT4D-H, SHREC'07-H}. 
Regarding partial shape matching, we include the well-known \textbf{SHREC'16} benchmark as well as partial-view datasets we construct based on \textbf{SCAPE\_r} and \textbf{FAUST\_r}. Besides, we further consider large-scale dataset -- \textbf{SURREAL} for training in the Supp. Mat..
	\item Animal: We utilize \textbf{TOSCA} dataset, which comprises various animal species.
We also consider the training of large-scale datasets from \textbf{SMAL} in the Supp. Mat..
	\item Garment:  We consider the \textbf{GarmCap}~\cite{lin2023leveraging} dataset includes four different garments, each is presented by a sequence of point clouds in various poses.
	\item Medical: We consider two datasets, \textbf{Spleen} and \textbf{Pancreas}, proposed in~\cite{adams2023point2ssm}. Both datasets are of limited data samples with considerable variabilities, which is typical in medical data processing. 
	\item Real-scans: The \textbf{Panoptic} dataset~\cite{joo2015panoptic} consists of partial point clouds derived from multi-view RGB-D images. We randomly select a subset of these views to recover partial point clouds.
\end{enumerate}

\noindent\textbf{Baseline:} We compare our method with a set of competitive baselines, including learning-based methods that can both train and test on point clouds -- CorrNet3D\cite{zeng2021corrnet3d}, RMA-Net~\cite{feng2021recurrent}, SyNoRiM \cite{huang_multiway_2022}, DiffFMaps~\cite{marin2020correspondence}, DPC \cite{lang2021dpc}, SE-ORNet\cite{deng2023se}, HSTR~\cite{he2023hierarchical} ; 
 methods required mesh for geometry-based training but inference with point clouds -- 3D-CODED~\cite{groueix20183d}, TransMatch~\cite{trappolini2021shape}, NIE\cite{nie}, SSMSM\cite{cao2023}, ConsistFMaps\cite{cao2022}, DPFM\cite{attaiki2021dpfm}, HCLV2S\cite{huang20};
non-learning-based axiomatic methods used for registration -- NDP~\cite{li2022non}   , AMM~\cite{amm}, PointSetReg~\cite{zhao2024clustereg}. 
 In Sec.~\ref{sec:realapp}, we further consider classical baselines especially tailored for medical dataset -- CAFE~\cite{cheng2021learning}, ISR~\cite{chen2020unsupervised}, Point2SSM~\cite{adams2023point2ssm}.
 Methods are marked according to if they need correspondence label ([S]) or not ([U]), as well as axiomatic ([A]).

\noindent\textbf{Remark: }In particular, we consider DPC~\cite{lang2021dpc} and SE-ORNet~\cite{deng2023se} as the primary competing methods because they can be trained purely on point clouds and without any correspondence annotation. 
We emphasize that meshing is generally non-trivial in real-world data (due to, \emph{e.g., }partiality and noise). 
The results of methods that require meshes during training are also included as \emph{reference}. 

\noindent\textbf{Evaluation metric:} Though we focus on the matching of point clouds, we primarily employ the widely-accepted geodesic error normalized by the square root of the total area of the mesh, to evaluate the performance of all methods.


\noindent\textbf{Hyper-parameters:} In Eqn.~\ref{eq:loss}, the hyper-parameters $\lambda_{\mbox{deform}},\lambda_{\mbox{arap}},\lambda_{\mbox{smooth}},\lambda_{\mbox{geo}}$ are uniformly set to 0.05, 0.005, 0.5, and 0.02 respectively. 
Model training utilizes the AdamW~\cite{loshchilov2017decoupled} optimizer with $\beta=(0.9,0.99)$, learning rate of 2e-3, and batch size of 2. 
We provide more details of hyper-parameters in the Supp. Mat.

\subsection{Standard non-rigid matching benchmarks}

In the following, we denote by $A/B$ the scheme of training on dataset $A$ and test on $B$. 

\noindent\textbf{Near-isometric benchmarks:} As illustrated in Tab.~\ref{table:iso_noiso}, our method consistently outperforms other purely point-based methods in all settings. 
Especially, our method achieves a promising performance improvement of over \textbf{59\%} compared to the previous SOTA approaches (\textbf{7.2} vs. 23.6; \textbf{9.3} vs. 23.0) in SCAPE\_r/SHREC19'\_r case and FAUST\_r/SHREC19'\_r case. 
Many previous methods performs well in the standard seen datasets but generalizes poorly to unseen shapes. 
Remarkably, our method outperforms all baselines, including the STOA method leveraging meshes during training, SSMSM \cite{cao2023}, in SCAPE\_r/FAUST\_r and FAUST\_r/SCAPE\_r (\textbf{5.1} vs. 8.5, \textbf{10.4} vs. 11.0).

\noindent\textbf{Generalization to non-isometric benchmarks:} We perform stress test on challenging non-isometric datasets including SHREC'07-H and DT4D-H. 
Our method achieves the best performance among \emph{all} methods as shown in Tab.~\ref{table:iso_noiso}, which indicates the excellent generalization ability to some unseen challenging cases. 
Especially, SHREC'07-H dataset comprises 20 heterogeneous human shapes with vertex numbers ranging from $3000$ to $15000$ and includes topological noise. 
Our method achieves a performance improvement of over \textbf{42\%} compared to the second best approach (\textbf{7.7} vs. 13.4; \textbf{8.2} vs. 15.3).

\noindent\textbf{Partial matching benchmarks:} 
As shown in Sec.~\ref{sec:loss}, our framework can be easily adapted for unsupervised partial-to-full point matching. 
We evaluate our method in two types of partial shape matching, including the challenging SHREC’16\cite{cosmo2016shrec} Cuts and Holes benchmark and two partial-view benchmarks built on SCAPE\_r and FAUST\_r datasets by ourselves, where we employ raycasting from the center of each face of a regular dodecahedron to observe the shapes, resulting $12$ partial view point clouds.

\begin{figure*}
  \centering
  \begin{subfigure}{0.4\linewidth}
    \includegraphics[width=9cm]{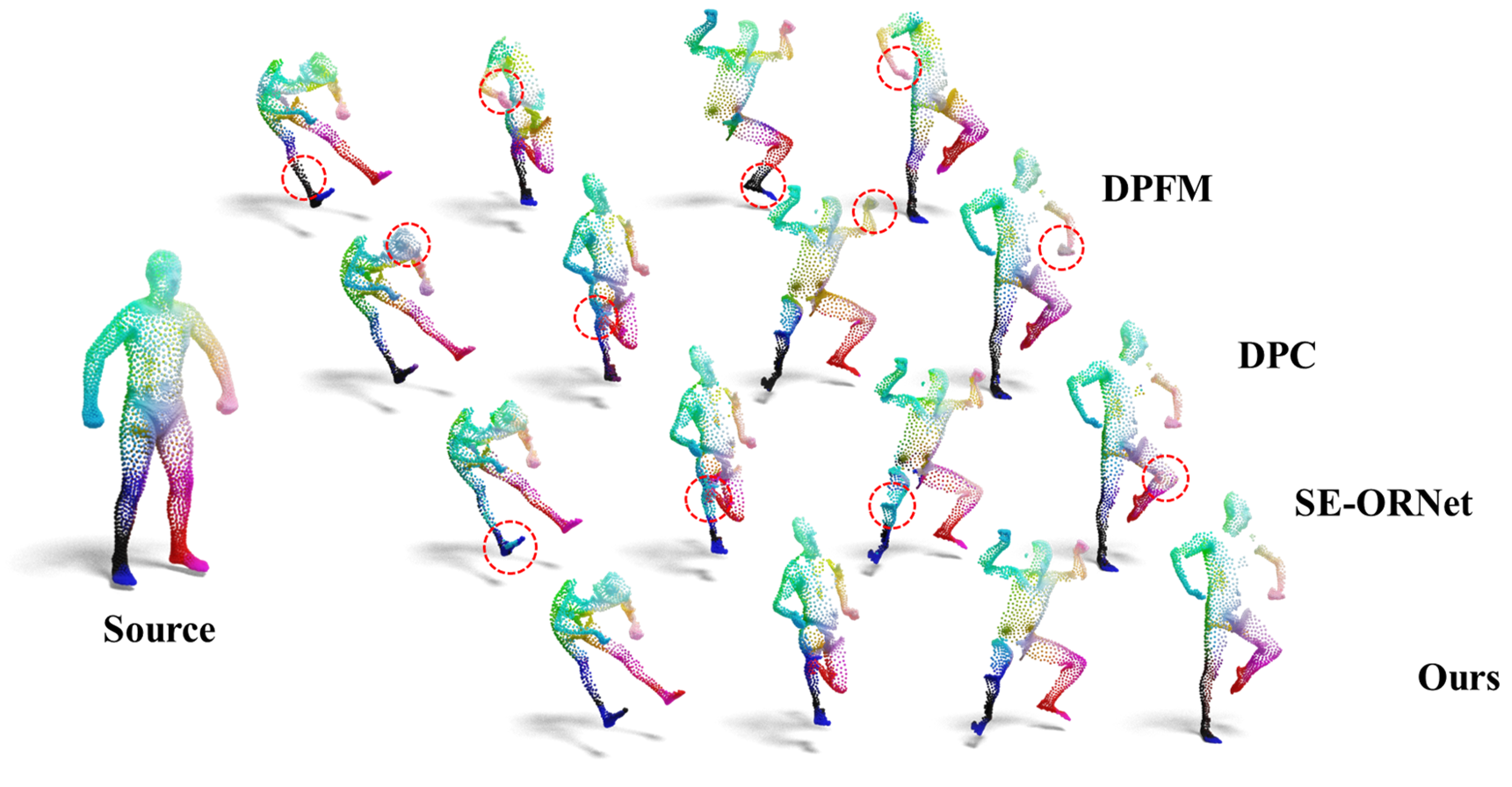}
    \caption{Partial view of SCPAE.}
    \label{fig:scape-pv}
  \end{subfigure}
  \hfill
  \begin{subfigure}{0.5\linewidth}
    \includegraphics[width=9cm]{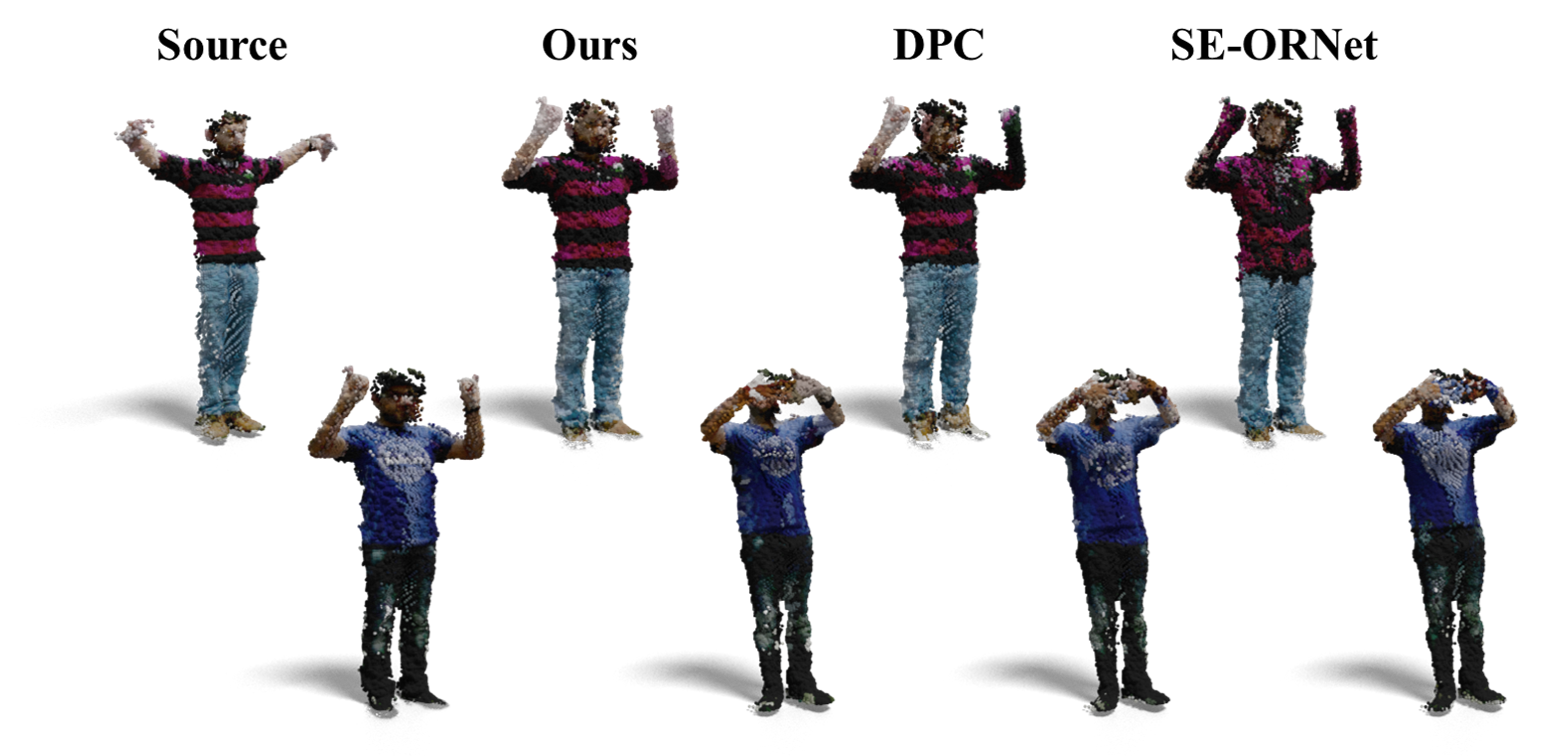}
    \caption{Noisy partial real scans.}
    \label{fig:realscan}
  \end{subfigure}
  \caption{Qualitative results of SCAPE-PV and noisy real scans.}
  \label{fig:pv}\vspace{-1.5em}
\end{figure*}

As illustrated in Tab.~\ref{table:partial}, our method outperforms the recent unsupervised method SSMSM~\cite{cao2023} in 3 out of 4 test cases, which requires meshes for training. 
Fig. \ref{fig:scape-pv} further shows qualitatively that our framework outperforms the competing methods including DPFM~\cite{attaiki2021dpfm}, which is based on mesh input as well. Moreover, recent axiomatic method PointSetReg~\cite{zhao2024clustereg} also struggles with partial cases.

Regarding purely point cloud-based baselines, we modify the reconstruction loss of~\cite{lang2021dpc, deng2023se} to unilateral loss to adapt partiality in the same way. 
In the end, we achieve the SOTA compared to them, exhibiting a significant over \textbf{54\%} superiority in partial view matching (\textbf{6.2} vs. 13.6; \textbf{5.3} vs. 13.9) and over \textbf{48\%} superiority in cuts/holes setting (\textbf{16.9} vs. 32.9; \textbf{13.0} vs. 27.6).

\begin{table}[t!]
    \caption{Quantitative results on partial cases including SCAPE-PV (S-PV), FAUST-PV(F-PV) and SHREC'16 (S16) in terms of mean geodesic errors $(\times100)$. 
    * indicates its original checkpoint using SURREAL190K. 
    The \textbf{best} results from the pure point cloud methods in each column are highlighted.}\label{table:partial}
    \vspace{-0.8em}
    \setlength{\tabcolsep}{0.7mm}
    \centering
    \scriptsize
    \begin{tabular}{lccclclc}
    \hline
    \multicolumn{1}{c}{}           &\textbf{Train}                                                                     & \multicolumn{2}{c}{\textbf{S-PV}} &  & \textbf{S16-CUTS}  &  & \textbf{S16-HOLES} \\ \cline{3-4} \cline{6-6} \cline{8-8} 
    \multicolumn{1}{c}{\multirow{-2}{*}{\textbf{Method}}} & \textbf{Test}       & \textbf{S-PV}      & \textbf{F-PV}     &  & \textbf{CUTS}          &  & \textbf{HOLES}         \\ \hline
    ConsistFMaps unsup{[}U{]}~\cite{cao2022}                            &                                              & \textbackslash{}              & \textbackslash{}             &  & 26.6          &  & 27.0          \\
    DPFM unsup{[}U{]}~\cite{attaiki2021dpfm}                                   &                                              & 11.5          & 15.2         &  & 20.9          &  & 22.8          \\
    HCLV2S*{[}S{]}~\cite{huang20}                                      &                                              & 8.7           & 5.3          &  & \textbackslash{}              &  & \textbackslash{}              \\
    SSMSM{[}U{]}~\cite{cao2023}                                        & \multirow{-4}{*}{\begin{tabular}[c]{@{}c@{}}Mesh\\ Required\end{tabular}}              & 8.8           & 8.0          &  & 12.2          &  & 16.7          \\
    \hline
    PointSetReg{[}A{]}~\cite{zhao2024clustereg}                                         &  & 15.7          & \textbackslash{}         &  & 36.1          &  & 24.0        \\
    DPC{[}U{]}~\cite{lang2021dpc}                                           &                                              & 13.6          & 14.5         &  & 32.9          &  & 32.5          \\
    SE-ORNet{[}U{]}~\cite{deng2023se}                                      & \multirow{-3}{*}{PCD Only}                   & 15.4          & 13.9         &  & 40.5          &  & 27.6          \\
    \rowcolor[HTML]{E7E6E6} 
    Ours{[}U{]}                                          & \multicolumn{1}{l}{\cellcolor[HTML]{E7E6E6}} & \textbf{6.2}  & \textbf{5.3} &  & \textbf{16.9} &  & \textbf{13.0} \\ \hline
    \end{tabular}
    \vspace{-1.5em}
\end{table}


\begin{table}[t!]
    \caption{Generalization performance of the checkpoint trained on sampled point cloud with fixed 1024 points of SHREC'19. We test this checkpoint on the more dense original point cloud. The \textbf{best} is highlighted. See more details of this evaluation in the Supp. Mat.}\label{table:points}
    \vspace{-0.8em}
    \centering
    \footnotesize
    \begin{tabular}{lcc}
    \hline
    \textbf{Method}          & \textbf{SHREC’19 (1024)} & \textbf{SHREC’19 (Ori.)} \\ \hline
    DPC{[}U{]}~\cite{lang2021dpc}      & 5.6             & 6.1 (+8.93\%)   \\
    SE-ORNet{[}U{]}~\cite{deng2023se} & 5.1             & 5.9 (+15.69\%)  \\
    \rowcolor[HTML]{E7E6E6} 
    Ours {[}U{]}     & \textbf{4.31}            & \textbf{4.33} (+0.46\%)  \\
    \hline
    \end{tabular}
    \vspace{-1.5em}
\end{table}

Finally, we attribute the above success to that the per-point features aggregated from pre-trained vision model which carries rich semantic information, helps to identify correspondence at the coarse level. 
In addition to that, our final features are further boosted by the geometric losses, leading to strong performance.

\subsection{Realworld Applications}\label{sec:realapp}

In this part, we showcase the utility of our framework under more practical settings: 

\begin{table}[]
\caption{Generalization testing on partial, isometric, and non-isometric full shape in terms of mean geodesic errors (×100).}\label{table:shrec}
\vspace{-0.8em}
\setlength{\tabcolsep}{1.0mm}
\centering
\footnotesize
\begin{tabular}{lcccc}
\hline
Method & SHREC19-PV    & SHREC19\_r    & DT4D-H        & SHREC07-H     \\ \hline
DPC~\cite{lang2021dpc}               & 32.77         & 32.65         & 23.71         & 18.94         \\
SE-ORNET~\cite{deng2023se}          & 29.18         & 28.06         & 18.61         & 29.99         \\
\rowcolor[HTML]{E7E6E6} 
Ours              & \textbf{9.40} & \textbf{8.97} & \textbf{8.61} & \textbf{9.50} \\ \hline
\end{tabular}
\vspace{-1.5em}
\end{table}
\noindent\textbf{Learning from synthetic partial scans: }
Since in practice, raw point clouds are often acquired as partial-view scans, we synthesize a set of $516$ partial point clouds with shapes from the SHREC'19 dataset. 
For simplicity, we add one template full point cloud as the reference and train DV-Matcher, DPC and SE-ORNet on the $516$ pairs (\emph{i.e., }full vs. partial). 
Note here we do not use any correspondence annotation. 
Then we evaluate the trained models on 1) test set of the above synthetic partial data, including pairs formed by a full and a partial point cloud (both randomly sampled from SHREC'19); 2) test set of SHREC'19 dataset, which are all full point clouds; 3) test set of DT4D-H and SHREC'07-H. 
As shown in Tab.~\ref{table:shrec}, our framework delivers consistently the best scores in all test cases, outperforming the baselines by a significant margin (see Fig.~\ref{fig:teaser} as well).


\noindent\textbf{Matching real scans: }
As shown in Fig. \ref{fig:realscan}, we transfer texture from the source shape (left-most) to target via maps from ours, DPC~\cite{lang2021dpc}, SE-ORNet~\cite{deng2023se}. Our method demonstrates smoother texture transfer compared to baselines (see particularly facial details and strips in the T-shirt). 

\noindent\textbf{Statistical shape models (SSM) for medical data: }Following Point2SSM \cite{adams2023point2ssm}, we evaluate our method on the anatomical SSM tasks. 
We stick to the regarding experimental setting and report our score in the spleen subset. 
As shown in Tab.~\ref{table:med}, our method outperforms the second best over \textbf{29\%} relative error reduction (\textbf{2.3} vs. 3.4; \textbf{1.9} vs. 2.7).

\begin{figure}[!t]
    \centering
    \includegraphics[width=0.5\textwidth]{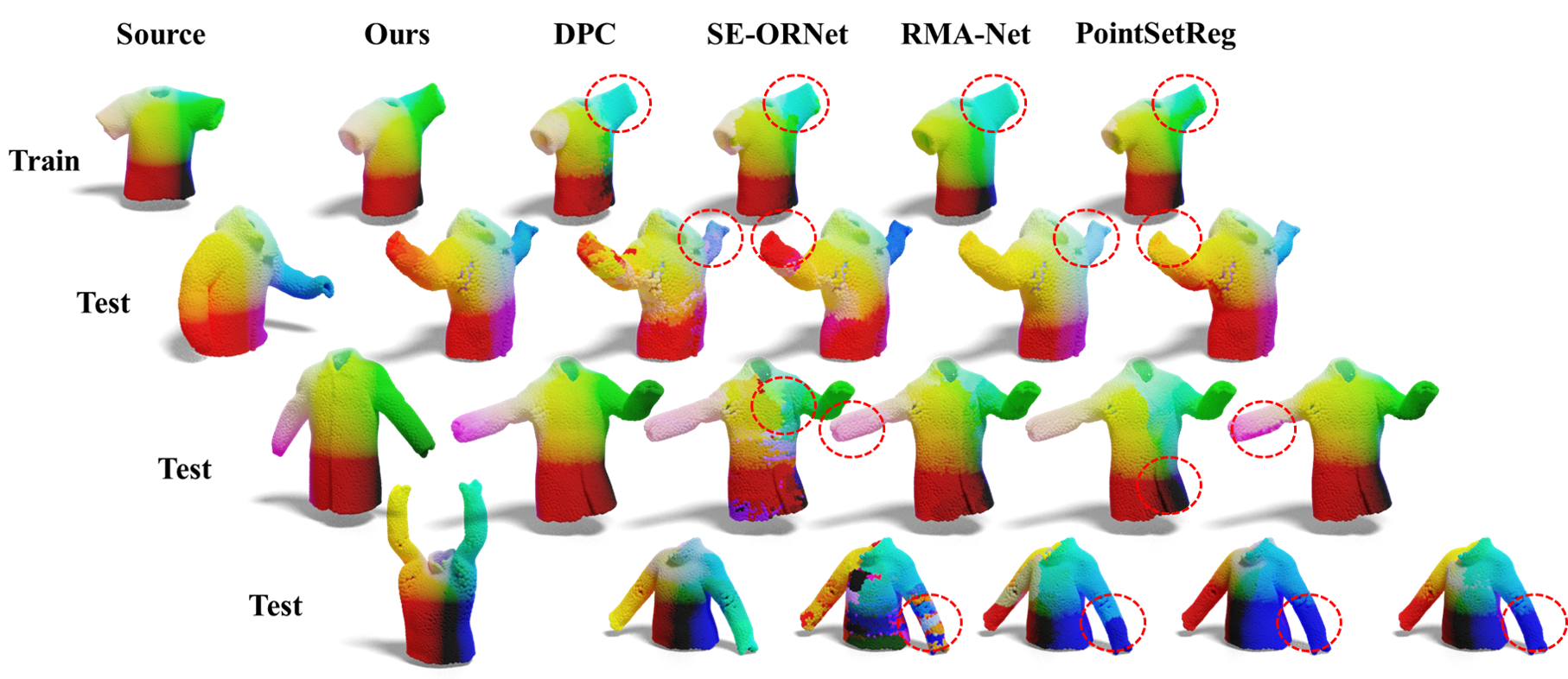}
    \caption{Qualitative results of GarmCap garment dataset.}\label{fig:cloth}
\end{figure}
\noindent\textbf{Garment dataset: } We in particular choose one of the four sequences of garment dataset~\cite{lin2023leveraging}, T-shirt, to train DV-Matcher and baselines. Then we directly evaluate on all test sets. 
As shown in Fig.~\ref{fig:cloth}, our method achieves the best accuracy as well as remarkable generalization performance, superior to the baselines, whether recent non-learning-based axiomatic methods ~\cite{zhao2024clustereg} or learning-based methods ~\cite{lang2021dpc},~\cite{deng2023se},~\cite{feng2021recurrent}.
We defer the quantitative results in the Supp. Mat. 

\begin{table}[t!]
    \caption{Statistical shape analysis on medical dataset in terms of chamfer distance (CD) and earth mover's distance (EMD). The \textbf{best} is highlighted.}\label{table:med}
    \vspace{-0.8em}
    \centering
    \footnotesize
    \begin{tabular}{lcccc}
    \hline
    \textbf{Dataset}                     & \multicolumn{2}{c}{\textbf{Spleen}} & \multicolumn{2}{c}{\textbf{Pancreas}} \\
    \textbf{Point Accuracy Metrics (mm)} & CD               & EMD              & CD                & EMD               \\ \hline
    CPAE{[}U{]}~\cite{cheng2021learning}                                 & 61.3             & 4.0              & 18.8              & 2.6               \\
    ISR{[}U{]}~\cite{chen2020unsupervised}                                  & 17.6             & 2.9              & 7.4               & 1.8               \\
    DPC{[}U{]}~\cite{lang2021dpc}                                  & 10.6             & 2.1              & 6.1               & 1.7               \\
    Point2SSM{[}U{]}~\cite{adams2023point2ssm}                            & 3.4              & 1.5              & 2.7               & 1.4               \\
    \rowcolor[HTML]{E7E6E6} 
    Ours{[}U{]}                                 & \textbf{2.3}     & \textbf{1.1}     & \textbf{1.9}      & \textbf{1.0}      \\ \hline
    \end{tabular}
    \vspace{-1.5em}
\end{table}

\subsection{Robustness and ablation analysis:} 
\begin{figure*}[!htbp]
    \centering
    \includegraphics[width=1\textwidth]{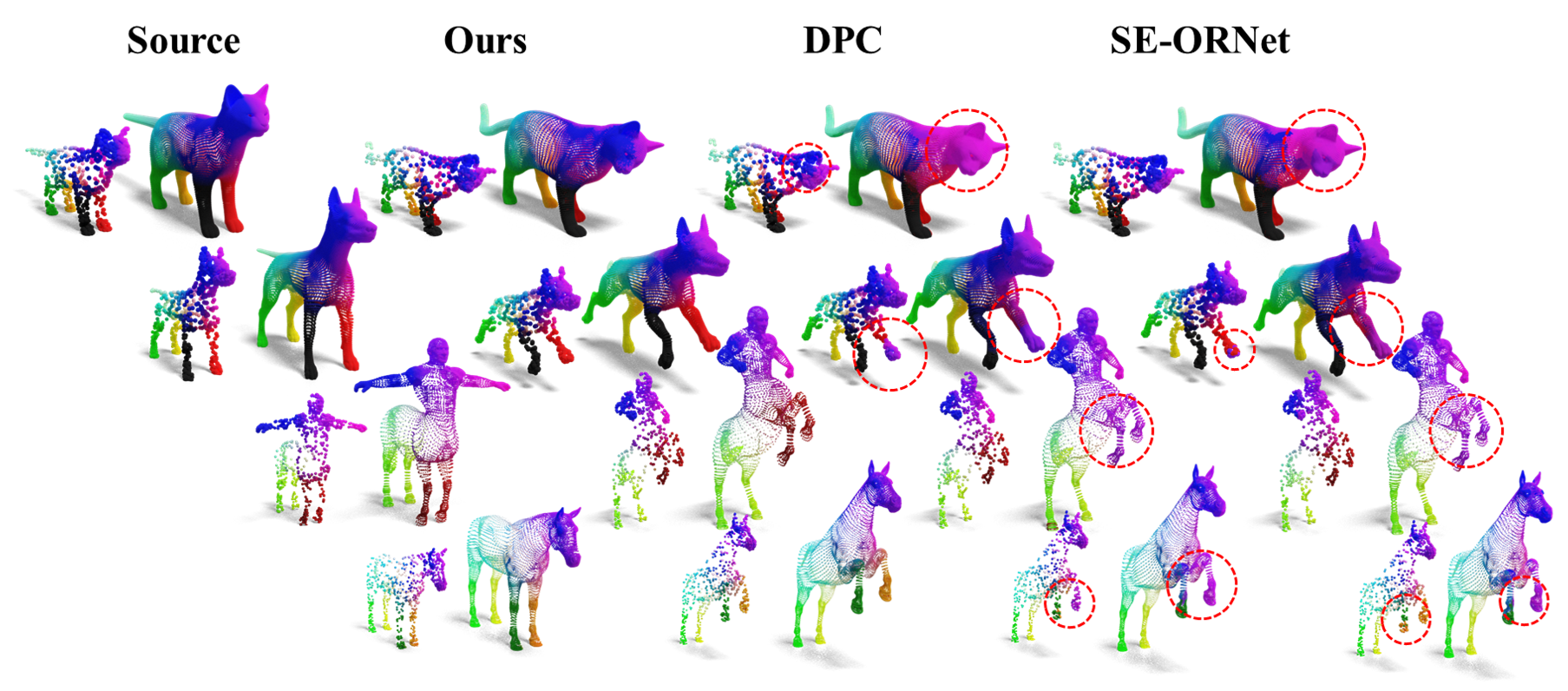}
    \vspace{-2em}
    \caption{Qualitative results of TOSCA. Our method demonstrates enhanced generalization capabilities when transitioning from sparse point clouds in training to dense point clouds in testing.}\label{fig:tosca}
    \vspace{-1.5em}
\end{figure*}
\noindent\textbf{Robustness analysis: }Reconstruction-based methods~\cite{lang2021dpc,deng2023se,he2023hierarchical} typically perform down-sampling to $n=1024$ points for \emph{both} training and testing. 
However, point clouds scanned in reality typically consist of tens of thousands of points, which is much denser.
To evaluate the robustness regarding point density, we use the checkpoint trained on down-sampled data released by the regarding authors to evaluate performance in both down-sampled test data ($1024$ points) and original test data ($\sim 5000$ points). 
As shown in Tab.~\ref{table:points}, DPC and SE-ORNet~\cite{lang2021dpc,deng2023se} both experience a degradation more than \textbf{8\%}. 
On the other hand, our method only yields a $0.46\%$ drop and achieves the best performance in both cases. 
Beyond the quantitative results, we also report qualitatively the generalization performance on TOSCA benchmark following the same setting as above, see Fig. \ref{fig:tosca} for more details. 

We attribute the above robustness to our introduction of pre-trained vision model in point feature learning, which effectively compensates for the discrepancy of low-resolution geometry.
Finally, we highlight that we have also performed robustness evaluation regarding noisy data and rotation perturbations in Supp. Mat.
\begin{table}[t!]
    \caption{Mean geodesic errors $(\times100)$ on different ablated settings, the models are all trained on SCAPE\_r and test on SCAPE\_r.}\label{table:ablation}
    \vspace{-1em}
    \centering
    \scriptsize
    \setlength{\tabcolsep}{0.5mm}
    \begin{tabular}{cccccc}
    \hline
    Ours-Recon & w/o visual enc.   & w/o $\mathcal{L}_{\mbox{deform}}$ & w/o $\mathcal{L}_{\mbox{geo}}$ & w/o $\mathcal{L}_{\mbox{smooth}}$ & w/o LG-Net \\
    7.6        & 15.1       & 11.2            & 12.8             & 6.9              & 17.3           \\ \hline
    w/o FeatUp & w/o LA-Net & w/o GA-Net      & w/o PE           & w/o Fusion      & Ours-Full      \\
    7.8        & 8.0        & 8.1             & 7.1              & 7.9              & 6.2            \\ \hline
    \end{tabular}\vspace{-2.5em}
\end{table}

\noindent\textbf{Ablation study: }
We first justify our overall design in Tab.~\ref{table:ablation}, where we sequentially remove each building block from our pipeline and train/test model on SCAPE\_r, including visual encoding (visual enc.) for pre-trained semantic priors, positional encoding (PE) for fine-grained positional information of each point, local and global attention network (LG-Net) for feature refining, etc. Specifically, the deformation-based loss we proposed plays a crucial role in efficient registration as illustrated in Fig.\ref{fig:abl}, as [with recon loss] would cause shape collapse, which delivers the difference between the methods simply using reconstruction loss like DPC~\cite{lang2021dpc}, SE-ORNET~\cite{deng2023se}, etc. Besides, our geometric loss ensures the preservation of local isometry as the [w/o geo loss] indicates. Furthermore, [w/o visual enc.] shows that our method is in synergy with the coarse-grained semantic information from pre-trained vision models. We have also assessed the necessity of each specific module within the LG-Net network in Tab.~\ref{table:ablation}, i.e., local attention blocks branch (LA-Net), global attention blocks branch (GA-Net), fusion module (Fusion) after the dual-pathway network.

\section{Conclusion and Limitation}
In this paper, we propose DV-Matcher for non-rigid point cloud matching that can be trained purely on point clouds without any correspondence annotation and extends naturally to partial-to-full matching. 
By incorporating semantic features from pre-trained vision models and a deformation-based proxy task, DV-Matcher demonstrates strong matching performance, promising generalizability and various robustness regarding partiality, point density, input orientation. 
Last but not least, DV-Matcher also performs well in real-world data, including 3D medical scans, textured garment scans, and noisy dynamic human scans. 

\begin{table}[]
\caption{Average time cost for each shape regarding SCAPE\_r.}\label{table:time}
\vspace{-0.8em}
\setlength{\tabcolsep}{1.5mm}
\centering
\footnotesize
\begin{tabular}{cccccc}
\hline
Method         & PointSetReg  & DFR                  & DPC                  & SE-ORNET             & \cellcolor[HTML]{E7E6E6}Ours          \\ \hline 
Time cost(s)         & 62.1[CPU]                  & 13.4                 & 1.41                 & 0.85        & \cellcolor[HTML]{E7E6E6}3.22 \\ \hline
\end{tabular}
\vspace{-1.5em}
\end{table}
\noindent\textbf{Limitation \& Future Work} 
While demonstrating superior performance and robustness over the baselines in a wide range of tests. 
We recognize the following limitations, which naturally give rise to future directions: 
1) As shown in Tab.~\ref{table:time}, our method takes on average $3.22$ seconds to match input of around $5000$ points, which is far from real-time; 
2) Though our method has shown robustness regarding certain rotation perturbations on inputs (see the Supp. Mat. for more details), it does not guarantee robustness in general orientation. In the future, we plan to further explore the ability of pre-trained visual models to address this limitation; 
3) Though our method performs the best in the stress test, it is still difficult to directly train on all raw scans, which requires overlapping region prediction. We believe it would be interesting to explore for higher practical impact. 
{
    \small
    \bibliographystyle{ieeenat_fullname}
    \bibliography{main}
}

\clearpage
\newpage
\appendix
\label{sec:appendix}
In this supplementary material, we provide more technical details and experimental results, including 1) A detailed description of our Visual Encoding, LG-Network in Sec.~\ref{sec:tech}; 2)  Detailed descriptions of datasets in Sec.~\ref{sec:data}; 3) Further qualitative results on matching heterogeneous shapes from SHREC’07-H and DT4D-H, quadruped shapes from SHREC'07-Fourleg and SHREC'20 in Sec.~\ref{sec:add1}, as well as the full/partial registration results; 4) Quantitative results following the setting from \cite{lang2021dpc,deng2023se,he2023hierarchical}, where train/test with the sparse point clouds of fixed $1024$ points in Sec.~\ref{sec:add2}; Besides, the qualitative result of garment datasets~\cite{lin2023leveraging} and SHREC'07-Fourleg are also presented in Sec.~\ref{sec:add2}; 5) Robustness evaluation of our method with respect to several perturbations on input in Sec.~\ref{sec:top}; 6) More high-dimensional feature visualization and matching results of different dataset in Sec.~\ref{sec:performance}; 7) Experimental setup, hyper-parameter instruction in Sec.~\ref{sec:env} and Sec.~\ref{sec:hyper} respectively. Finally, the broader impacts are discussed in Sec.~\ref{sec:bro}.

\section{Technical Details}\label{sec:tech}
\begin{figure}[!h]
    \centering
    \includegraphics[width=0.45\textwidth]{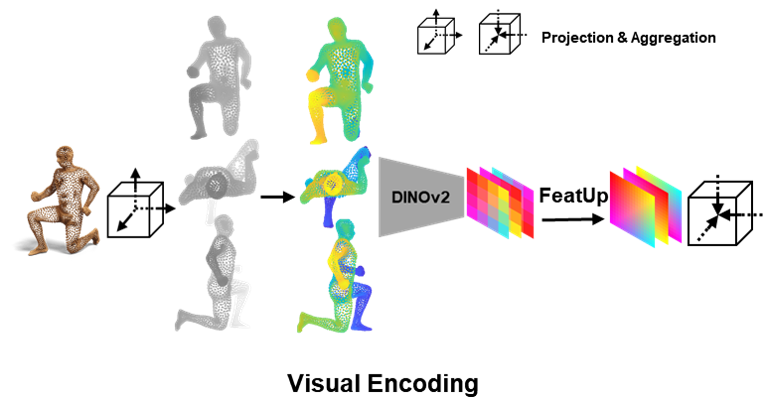}
    \caption{The schematic illustration of the proposed Visual Encoding.}\label{fig:visual_encoding}
\end{figure}
\begin{figure}[!t]
    \centering
    \includegraphics[width=0.47\textwidth]{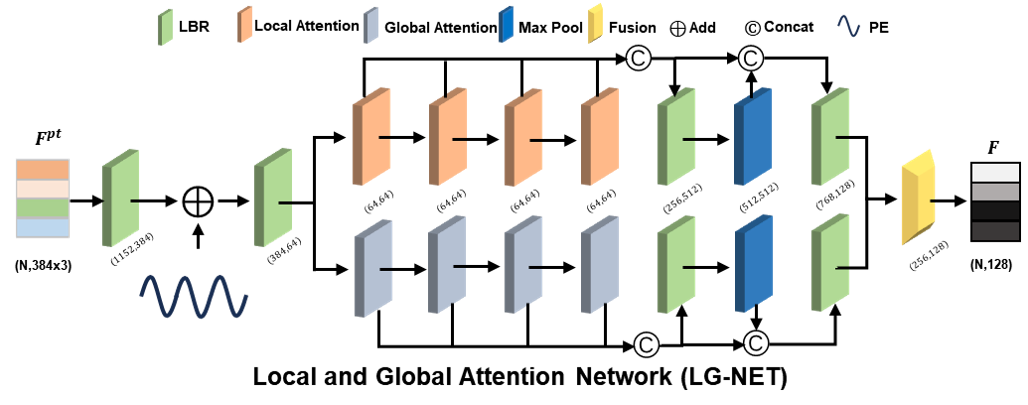}
    \caption{The schematic illustration of LG-Net.}\label{fig:lgnet}
\end{figure}
\noindent\textbf{Visual Encoding:} 
Fig.~\ref{fig:visual_encoding} elucidates how we leverage the utilization of features from pre-trained vision models through a point-wise invertible projection from 3D point clouds to 2D images. Specifically, we get features through DINOv2~\cite{oquab2023dinov2} and lift features via FeatUp~\cite{fu2024featup}, where the semantic features are then back-projected to their corresponding points, as described in Sec.~\label{sec:per-point} of the main text.

\noindent\textbf{LG-Net:} 
Fig.~\ref{fig:lgnet} illustrates the composition of LG-Net, which aims to refine the features learned from 2D pre-trained vision models, so that is robust to large deformations and generalized to the challenging partiality. Specifically, for the input representation $F^{pt}(P)$ derived from pre-trained vision models, we employ the LBR~\cite{guo2021pct}, which combines Linear, BatchNorm, and ReLU layers, to facilitate the feature dimension transformation into $F_{\Theta}^{\prime}\in \mathbb{R} ^{N \times 384}$. Following this, we apply position encoding from ~\cite{mildenhall2021nerf} to integrate 3D absolute position information, which is subsequently combined with the block-wise semantic features $F_{\Theta}^{\prime}$ to yield a refined representation $F_{\Theta}\in \mathbb{R} ^{N  \times 384 }$, denotes $F_{\Theta}=F_{\Theta}^{\prime} + \gamma$, where $\gamma$ is a mapping from $\mathbb{R}$ into a higher dimensional space $\mathbb{R}^{N\times384}$. Later, our designed network is a dual-pathway architecture in parallel to refine $F_{\Theta}$, comprising \emph{Global Attention} and \emph{Local Attention}. The two attention modules differ in receptive field -- given a point, the former abstracts features of the remaining points to achieve comprehensive global perceptual awareness, while the latter focuses on its nearest neighborhoods. After undergoing local and global attention mechanisms respectively, we further fuse both features at the end of refined network to obtain a more comprehensive feature representation. Where \emph{Fusion} module consists of LBR and a three-layer stacked N2P~\cite{wu2023attention} attention to merge features.

\begin{figure*}[!t]
    \centering
    \includegraphics[width=1\textwidth]{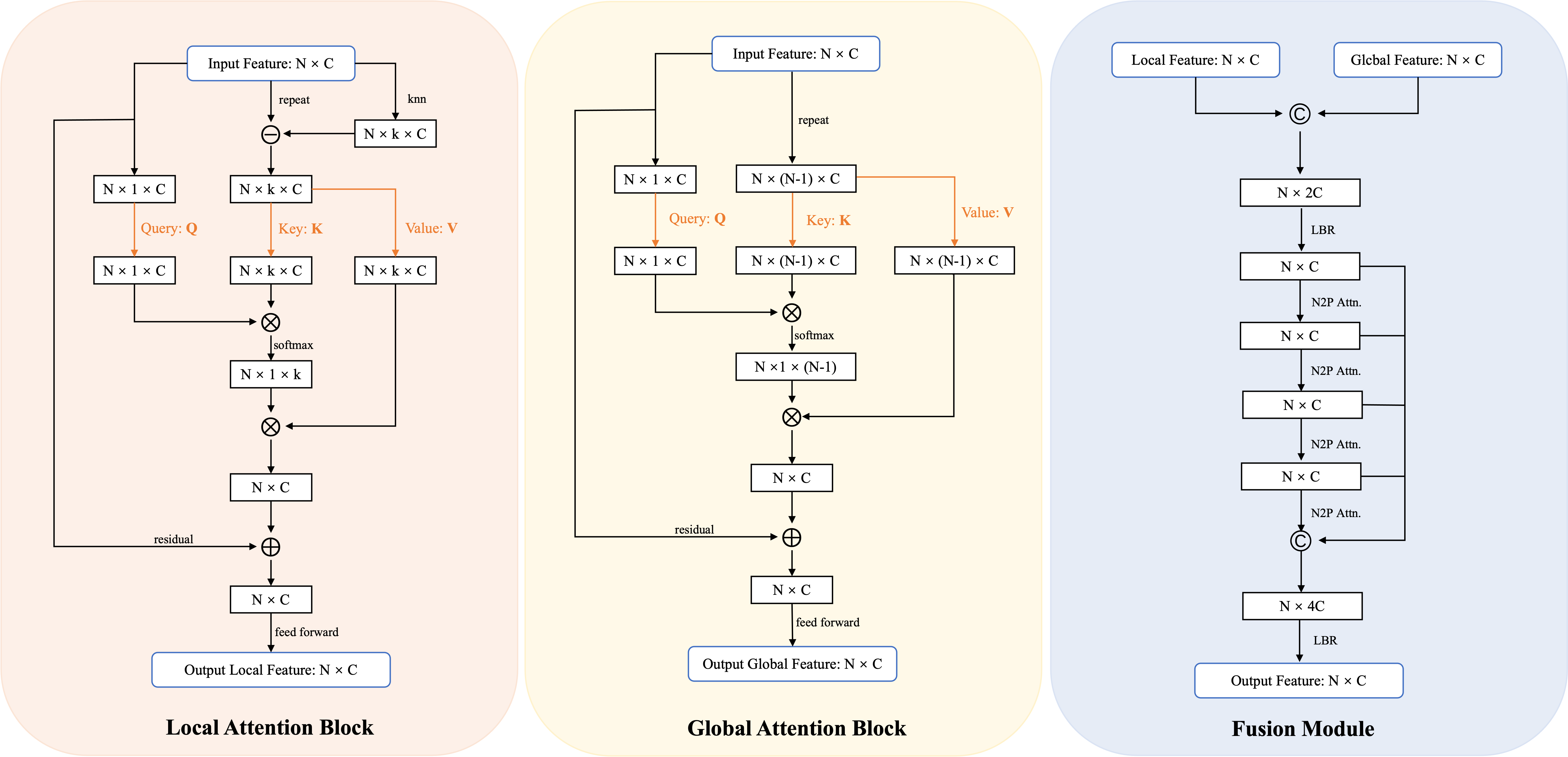}
    \caption{The schematic illustration of the main blocks of LG-Net.}\label{fig:network}
\end{figure*}
\noindent\textbf{Network Details:} 
Fig.~\ref{fig:network} depicts from left to right the architecture diagrams of our \emph{local attention block}, \emph{global attention block}, and \emph{fusion module}. 

\noindent\textbf{ARAP loss: } The as-rigid-as-possible term is also incorporated following \cite{guo2021human, levi2014smooth}, which reflects the deviation of estimated local surface deformations from rigid transformations:
\begin{equation}
\mathcal{L}^{(\src,\tar)}_{\mbox{arap}}=ARAP(\textbf{X})
\end{equation}
\begin{equation}
\footnotesize
d_{h, l}(\mathbf{X})=d_{h, l}(\Theta, \Delta)=R\left(\Theta_h\right)\left(g_l-g_h\right)+\Delta_k+g_k-\left(g_l+\Delta_l\right).
\end{equation}

Here, $g \in \mathbb{R}^{H \times 3}$ are the original positions of the nodes in the deformation graph $\mathcal{D} \mathcal{G}$, $\psi(h)$ denotes the 1-ring neighborhood of the $h-$ th deformation node. $R(\cdot)$ corresponds to Rodrigues' rotation formula, which computes a rotation matrix from an axis-angle representation, and $\alpha$ is the weight of the smooth rotation regularization term. 

\section{Dataset Details}\label{sec:data}
\textbf{SCAPE\_r:} The remeshed version of the SCAPE dataset~\cite{scape} comprises 71 human shapes. We split the first 51 shapes for training and the rest 20 shapes for testing.
\textbf{FAUST\_r:} The remeshed version of FAUST dataset \cite{bogo2014faust} comprises 100 human shapes. We split the first 80 shapes for training and the rest 20 for testing. 
\textbf{SHREC'19\_r:} The remehsed version of SHREC19 dataset~\cite{melzi2019shrec} comprises 44 shapes. We pair them into 430 annotated examples provided by \cite{melzi2019shrec} for testing. 
\textbf{DT4D-H:} A dataset from \cite{magnet2022smooth} comprises 10 categories of heterogeneous humanoid shapes. Following \cite{dfr}, we use it solely in testing, and evaluating the inter-class maps split in \cite{magnet2022smooth}. 
\textbf{SHREC'07-H:} A subset of SHREC'07 dataset \cite{giorgi2007shape} comprises 20 heterogeneous human shapes. We use it solely in testing. 
\textbf{SHREC'07-Fourleg:} A subset of SHREC'07 dataset \cite{giorgi2007shape} comprises 20 heterogeneous fourleg animals. We use a total of 380 pairs for training.
\textbf{SHREC'20:} A dataset~\cite{Dyke:2020:track.b} comprising highly non-isometric non-rigid quadruped shapes of 14 animals, encompassing 12 full shapes and 2 partial shapes. We use it solely for testing. 
\textbf{SURREAL:} It is the large-scale dataset from \cite{groueix20183d} comprises 230,000 training shapes, from which we take the first 2,000 shapes and use it solely for training. 
\textbf{TOSCA:} Dataset from \cite{zuffi20173d} comprises 41 different shapes of various animal species. Following \cite{lang2021dpc,deng2023se}, we pair these shapes to create both for training and evaluation, respectively. \textbf{SHREC'16:} Partial shape dataset SHREC'16 \cite{cosmo2016shrec} includes two subsets, namely CUTS with 120 pairs and HOLES with 80 pairs. Following \cite{attaiki2021dpfm,cao2023}, we train our method for each subset individually and evaluate it on the corresponding unseen test set (200 shapes for each subset). Moreover, we further conduct some practical experiments on partial real scan dataset processed from \cite{Joo_2017_TPAMI} and medical dataset from \cite{adams2023point2ssm}.
\textbf{SURREAL:} The large-scale dataset from \cite{groueix20183d} comprises 230,000 training shapes, from which we select the first 2,000 shapes and use them solely for training. 
\textbf{SMAL:} Large-scale dataset from~\cite{zuffi20173d}, which includes parameterized animal models for generating shapes. 
We employ the model to generate $2000$ instances of diverse poses for each animal category, resulting in a training dataset comprising $10000$ shapes. 
\textbf{GarmCap:} A dataset from~\cite{lin2023leveraging}, which contains textured 3D garment scans in various poses. We take 40 T-shirt shapes for training, and test on 10 unseen T-shirt shapes, along with 10 long coats, 10 thick coats, and 10 orange coats.
\textbf{Spleen:} Following~\cite{adams2023point2ssm}, we take 32 aligned medical spleens for training and 4 other shapes for testing.
\textbf{Pancreas:} Following~\cite{adams2023point2ssm}, we also take 216 aligned medical pancreases for training and 28 other shapes for testing.

\section{Additional Experiments}
\subsection{Further Qualitative Results }\label{sec:add1}
\begin{figure*}[!htbp]
    \centering
    \includegraphics[width=1\textwidth]{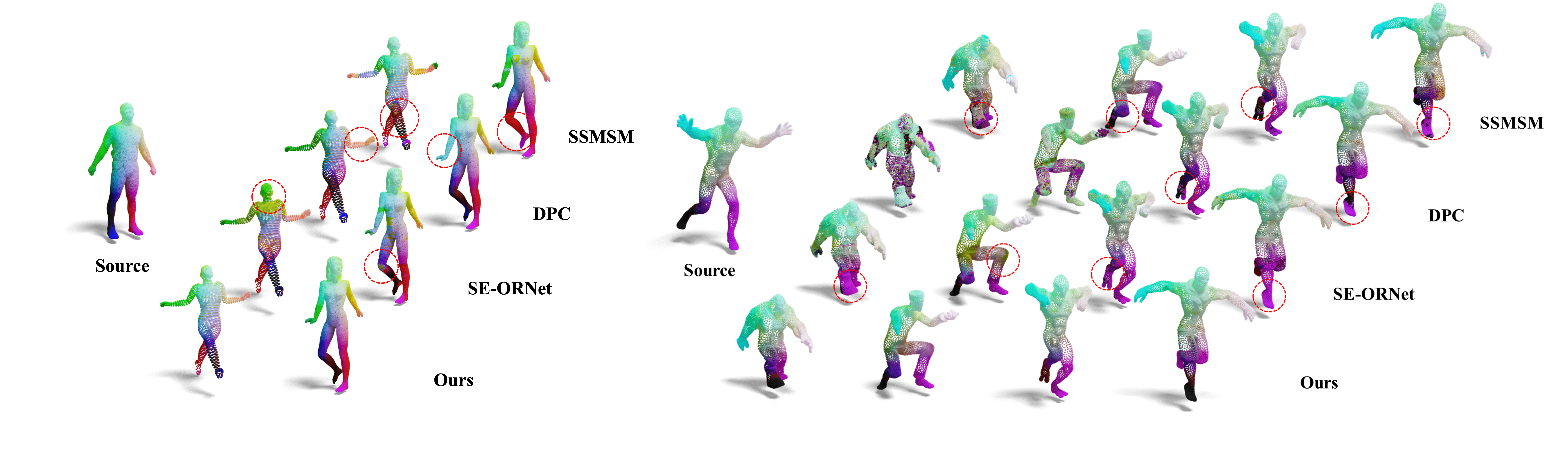}
    \caption{We estimate correspondences between heterogeneous shapes from SHREC’07-H and DT4D-H with DPC,SE-ORNET and one SSMSM, all trained on the SCAPE\_r dataset. Our method outperforms the competing methods by a large margin.}\label{fig:noiso}
\end{figure*}
\begin{figure*}[!htbp]
    \centering
    \includegraphics[width=1\textwidth]{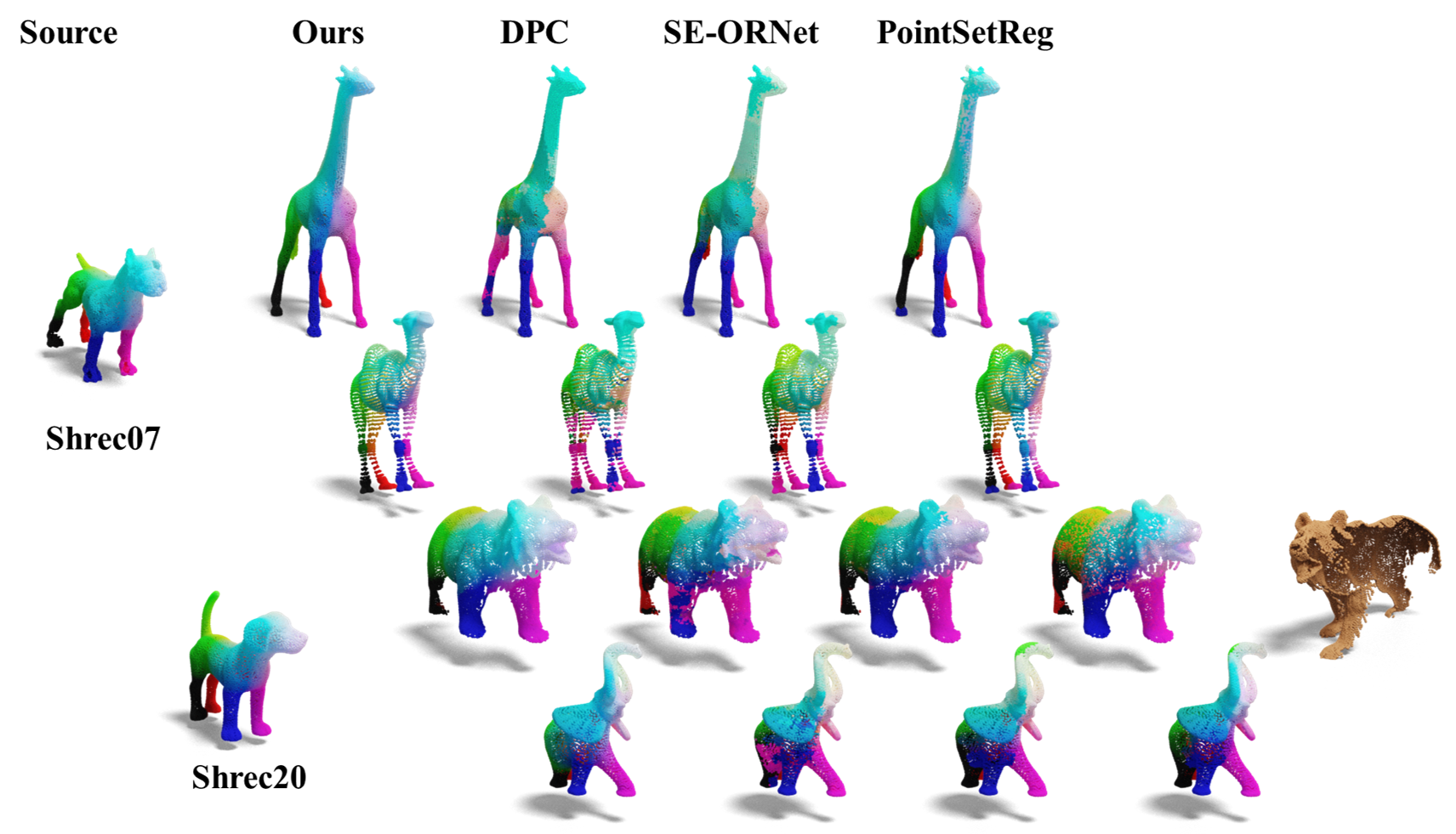}
    \caption{We estimate correspondences between highly non-isometric non-rigid quadruped shapes from SHREC’07-Fourleg and SHREC'20 with DPC,SE-ORNET and PointRegSet, all learning-based methods trained on the SHREC’07-Fourleg dataset. Our method outperforms the competing methods by a large margin. Note that the bear in the third row is incomplete. }\label{fig:shrec}
    \vspace{1.5em}
\end{figure*}
\begin{figure}[!t]
    \centering
    \includegraphics[width=0.47\textwidth]{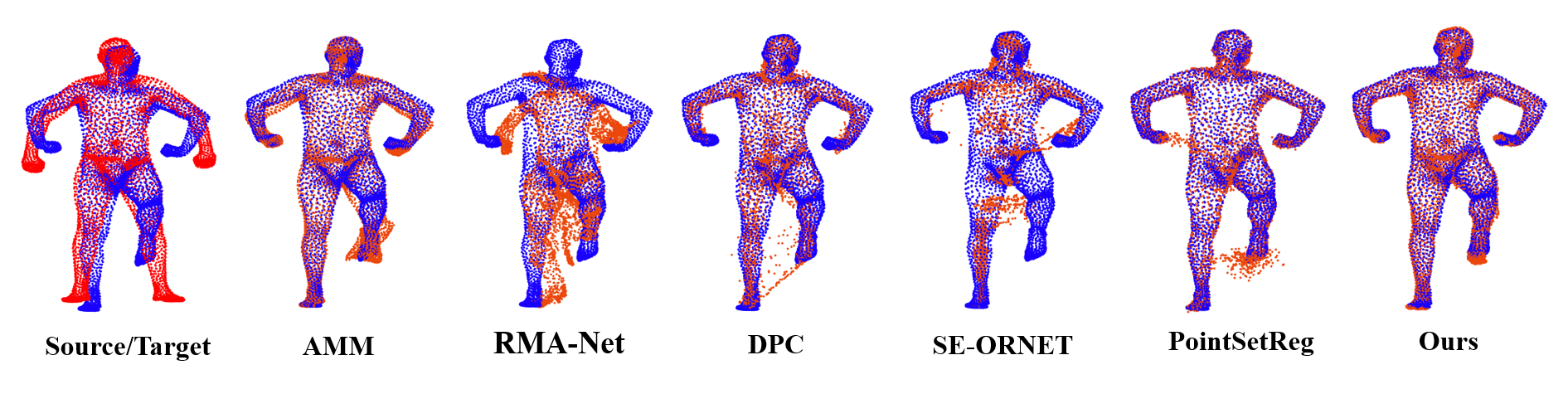}
    \vspace{-1em}
    \caption{The figure illustrates the registration results of various baselines, along with our proposed deformer.}\label{fig:reg}
\end{figure}
\begin{figure}[!t]
    \centering
    \includegraphics[width=0.47\textwidth]{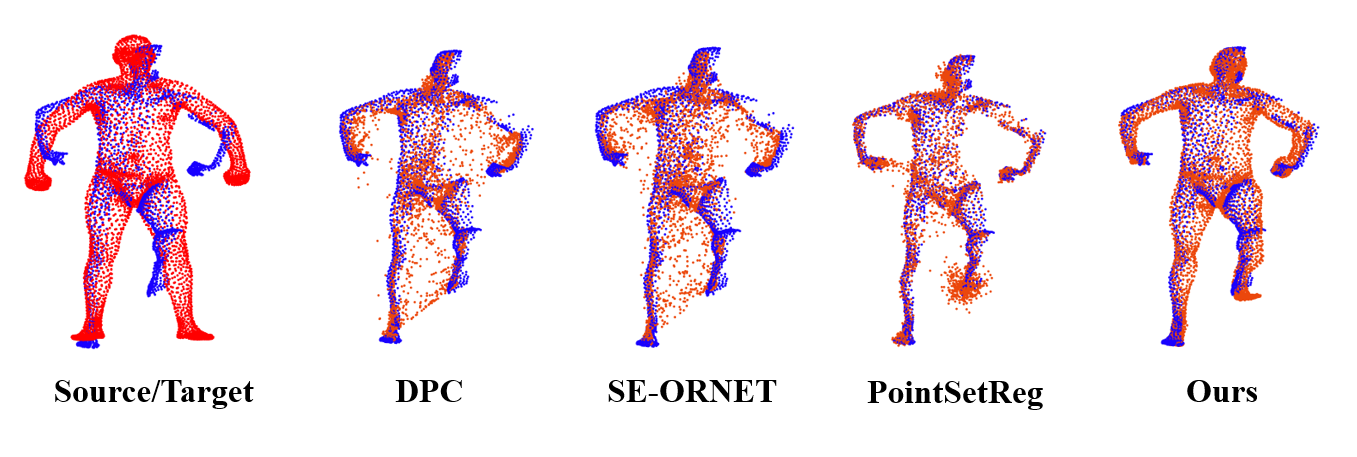}
    \vspace{-1em}
    \caption{The figure illustrates the partial registration results of various baselines, along with our proposed deformer.}\label{fig:reg_partial}
\end{figure}
\noindent\textbf{Non-isometric human shapes Matching:} In Fig.~\ref{fig:noiso}, we qualitatively visualize maps obtained by different methods tested in the SHREC’07-H and DT4D-H benchmark. 
It is obvious that our results outperform all the competing methods, showing superior generalization performance. 

\noindent\textbf{Non-isometric quadruped shapes Matching:} We also conducted training on the quadruped dataset -- SHREC'07-Fourleg, and subsequently tested on challenging SHREC'07 and SHREC'20, respectively. Fig.~\ref{fig:shrec} illustrates several of the most highly non-isometric shapes, where our method significantly outperforms other baselines. Specifically, as we leverage the semantic information extracted from pre-trained vision models and the formulation of geometric information, our approach exhibits promising performance across even challenging heterogeneous shapes.

\noindent\textbf{Full Registration Results:} Fig.~\ref{fig:reg} illustrates the registration results of different methods on full point clouds, where all learning-based methods were trained on SCAPE\_r dataset. The results indicate that axiomatic non-learning-based methods, whether AMM~\cite{amm} or the recent PointSetReg~\cite{zhao2024clustereg}, all exhibit errors in the vicinity of the foot area; whereas, learning-based reconstruction methods -- DPC~\cite{lang2021dpc}, SE-ORNet~\cite{deng2023se} reconstruct the point clouds with substantial noise; RMA-Net~\cite{feng2021recurrent}, which also employs projected 2D images as a prior, but fails to deform effectively to the target shape as well. In contrast, our deformer achieves efficient, high-quality, and smooth deformed point clouds quickly without optimizing iterations.

\noindent\textbf{Partial Registration Results:} Fig.~\ref{fig:reg_partial} presents more challenging cases, namely registering the full point cloud to the partial point cloud, where all learning-based methods were trained on SCAPE-PV dataset. The results show that all other baselines fail to maintain the complete source shape after registration, collapsing into partial, and both learning-based methods~\cite{lang2021dpc,deng2023se} and recent axiomatic non-learning-based method~\cite{zhao2024clustereg} result in significant noise post-registration. This further underscores the robustness of our method and our ability to handle partial cases effectively.

\subsection{Further Quantitative Results }\label{sec:add2}
For the benchmarks involving downsampled point clouds from original shapes, which results in the absence of the complete mesh structure. Thus, we replace the geodesic distance with the Euclidean distance for our evaluation, as defined in Eq.~\ref{eq:err}. This substitution is detailed in Tab.~\ref{table:baseline} and Tab.~\ref{table:cloth} in Supp. Mat., also Tab.~\ref{table:points} within the main text.

\noindent\textbf{Sparse Humans/Animals Benchmarks:} Following the prior works~\cite{lang2021dpc,deng2023se,he2023hierarchical}, we conduct the experiments with a consistent sampling point number of $n=1024$. Specifically, for a pair of source and target shapes $(\src, \tar)$, the correspondence error is defined as:
\begin{equation}
\label{eq:err}
err=\frac{1}{N} \sum_{x_i \in \src}\left\|f\left(x_i\right)-y_{g t}\right\|_2,
\end{equation}
where $y_{g t} \in \tar$ is the ground truth corresponding point to $x_i$. Additionally, we measure the correspondence accuracy, defined as:

\begin{equation}
acc(\epsilon)=\frac{1}{N} \sum_{x_i \in \src} \mathbb{I}\left(\left\|f\left(x_i\right)-y_{g t}\right\|_2<\epsilon d\right),
\end{equation}
where $\mathbb{I}(\cdot)$ is the indicator function, $d$ is the maximal Euclidean distance between points in $\tar$, and $\epsilon \in[0,1]$ is an error tolerance. We evaluate the accuracy at 1\% tolerance following~\cite{lang2021dpc}.

We train on the SURREAL and SHREC’19 dataset respectively, and then test on the SHREC’19 dataset. Similarly, we train respectively on SMAL and TOSCA dataset, and then test on the TOSCA dataset. 
As shown in Tab.~\ref{table:baseline}, unlike HSTR\cite{he2023hierarchical}, which achieves the best performance on its intra-dataset but lags behind SE-ORNet\cite{deng2023se} on cross-dataset generalization, our approach excels in both intra-dataset and cross-dataset tests, surpassing all existing methods by over \textbf{12\%} (\textbf{4.3} vs. 4.9). 
This also complements Tab.~\ref{table:iso_noiso} in the main text, demonstrating that our method yields robust results whether trained/tested on dense or sparse point clouds.
\begin{figure*}[!t]
    \centering
    \includegraphics[width=1\textwidth]{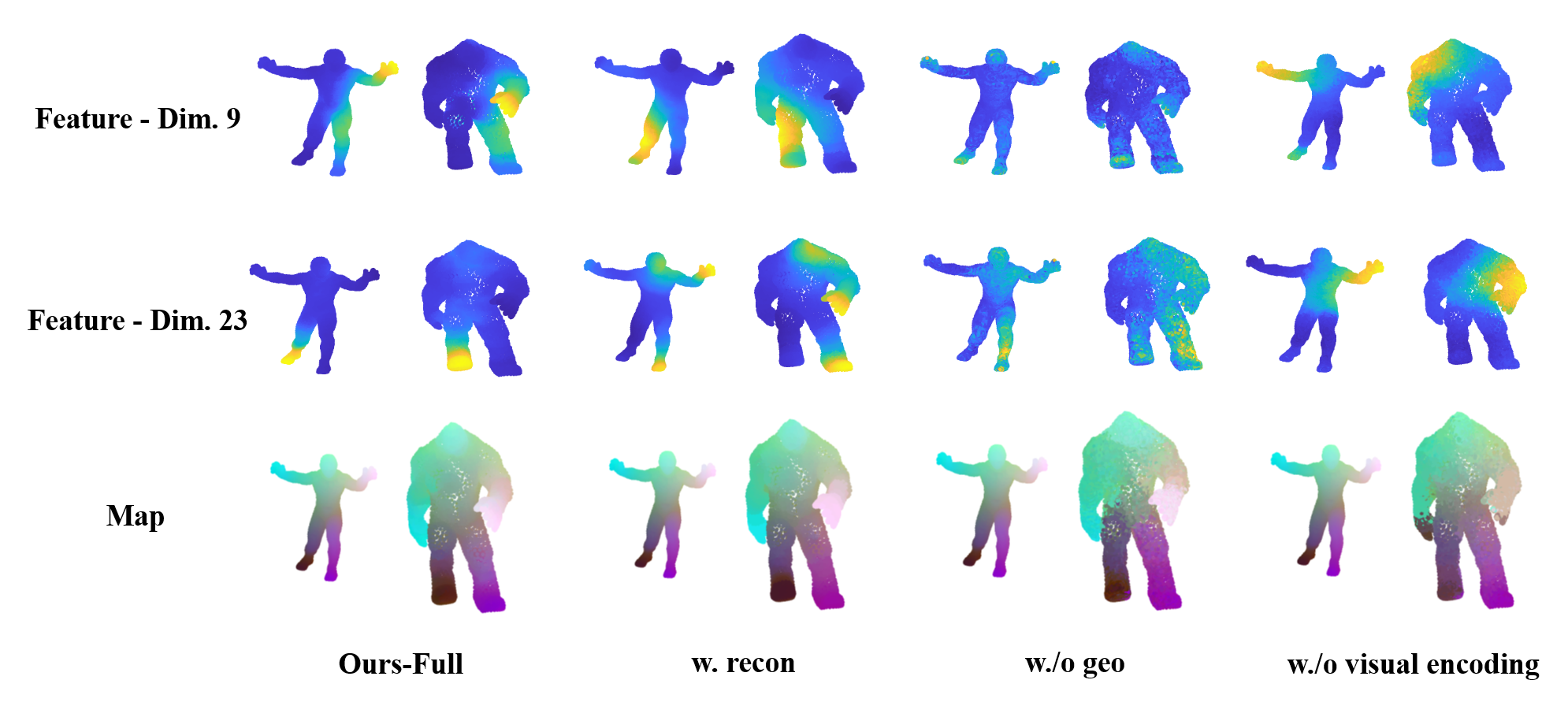}
    \vspace{-1.5em}
    \caption{Visualization of different feature dimensions and mapping. $\textbf{Dim.i}$ denotes the features of the $i-th$ dimension, where $i \leq 128$.
    }\label{fig:feature}
    \vspace{-0em}
\end{figure*}
\begin{table}[t!]
    \caption{Quantitative results on human and animals datasets. Acc signifies correspondence accuracy at 0.01 error tolerance, and err denotes average correspondence error ($err \times 1000$). The \textbf{best} results in each column are highlighted.}\label{table:baseline}
    \centering
    \scriptsize
    \setlength{\tabcolsep}{0.8mm}
    \begin{tabular}{lcccclcccc}
    \hline
    \multicolumn{1}{c}{\textbf{Train}} & \multicolumn{2}{c}{\textbf{SHREC’19}} & \multicolumn{2}{c}{\textbf{SURREAL}} &  & \multicolumn{2}{c}{\textbf{TOSCA}} & \multicolumn{2}{c}{\textbf{SMAL}} \\ \cline{2-5} \cline{7-10} 
    \multicolumn{1}{c}{\textbf{Test}}  & \multicolumn{4}{c}{\textbf{SHREC’19}}                                        &  & \multicolumn{4}{c}{\textbf{TOSCA}}                                     \\
                                       & acc ↑               & err ↓           & acc ↑              & err ↓           &  & acc ↑             & err ↓          & acc ↑             & err ↓         \\ \hline
    3D-CODED{[}S{]}~\cite{groueix20183d}                   & \textbackslash{}                     & \textbackslash{}                 & 2.1\%              & 8.1             &  & \textbackslash{}                   & \textbackslash{}                & 0.5\%             & 19.2          \\
    Elementary{[}S{]}~\cite{deprelle2019learning}                           & \textbackslash{}                     & \textbackslash{}                 & 0.5\%              & 13.7            &  & \textbackslash{}                   & \textbackslash{}                & 2.3\%             & 7.6           \\
    CorrNet3D{[}U{]}~\cite{zeng2021corrnet3d}                 & 0.4\%               & 33.8            & 6.0\%              & 6.9             &  & 0.3\%             & 32.7           & 5.3\%             & 9.8           \\
    RMA-Net{[}U{]}~\cite{feng2021recurrent}                 & 4.5\%               & 6.0            &  \textbackslash{}            &  \textbackslash{}            &  & 2.2\%             & 29.4           &  \textbackslash{}          & \textbackslash{}          \\
    DPC{[}U{]}~\cite{lang2021dpc}                        & 15.3\%              & 5.6             & 17.7\%             & 6.1             &  & 34.7\%            & 2.8            & 33.2\%            & 5.8           \\
    SE-ORNet{[}U{]}~\cite{deng2023se}                   & 17.5\%              & 5.1             & 21.5\%             & 4.6             &  & 38.3\%            & 2.7            & 36.4\%            & 3.9           \\
    HSTR{[}U{]}~\cite{he2023hierarchical}                      & 19.3\%              & 4.9             & 19.4\%             & 5.6             &  & 52.3\%   & 1.2            & 33.9\%            & 5.6           \\
    \rowcolor[HTML]{E7E6E6} 
    Ours {[}U{]}                       & \textbf{23.9\%}     & \textbf{4.3}    & \textbf{27.1\%}    & \textbf{4.0}    &  & \textbf{56.2\%}            & \textbf{0.9}   & \textbf{39.5\%}   & \textbf{3.3}  \\
    \hline
    \end{tabular}\vspace{-0em}
\end{table}

\noindent\textbf{Garment Dataset:} We choose T-shirt (GarmCap\_1) to train our DV-Matcher and other baselines, then evaluate on all four sequences of garment dataset~\cite{lin2023leveraging}. As shown in Tab.~\ref{table:cloth}, our method outperforms the second best over \textbf{35\%} relative error reduction (\textbf{5.24} vs. 8.09).

\noindent\textbf{SHREC'07-Fourleg Dataset:}
We conducted further validation on challenging heterogeneous quadrupeds. We selected all 20 shapes and uniformly sampled (including upsampling and downsampling) to 5,000 points for training, and tested on 380 pairs of original point clouds. As shown in Tab.~\ref{table:fourleg}, our method outperforms \textbf{49\%} over past approaches (\textbf{6.19} vs. 12.37), whether they are learning-based~\cite{lang2021dpc,deng2023se} or axiomatic~\cite{zhao2024clustereg}.

\begin{table}[]
\caption{Quantitative results on four different garments from GarmCap in terms of Euclidean distance error ($err \times 100$). The \textbf{best} is highlighted.}\label{table:cloth}
\vspace{-0.8em}
\setlength{\tabcolsep}{1mm}
\centering
\scriptsize
\begin{tabular}{lcccc}
\hline
\textbf{Method} & \textbf{GarmCap\_1} & \textbf{GarmCap\_2}  & \textbf{GarmCap\_3}  & \textbf{GarmCap\_4}  \\ \hline
PointSetReg{[}A{]}~\cite{zhao2024clustereg}         & 8.95          & 8.53          & 9.27          & 8.96          \\
DPC{[}U{]}~\cite{lang2021dpc}        & 7.24          & 10.12         & 10.19         & 9.03          \\
SE-ORNET{[}U{]}~\cite{deng2023se}    & 7.11          & 8.09          & 10.08         & 8.86          \\
RMA-Net{[}U{]}~\cite{feng2021recurrent}     & 7.02          & 10.92         & 9.55          & 9.67          \\
\rowcolor[HTML]{E7E6E6} 
Ours{[}U{]}                 & \textbf{4.92} & \textbf{5.24} & \textbf{5.85} & \textbf{5.62} \\ \hline
\end{tabular}
\end{table}
\begin{table}[!t]
\caption{Quantitative results on SHREC'07-Fourleg in terms of mean geodesic distance errors ($\times 100$). The \textbf{best} is highlighted.}\label{table:fourleg}
\centering
\scriptsize
\begin{tabular}{lc}
\hline
\textbf{Method} & \multicolumn{1}{l}{\textbf{SHREC'07-Fourleg}} \\ \hline
DPC{[}U{]} ~\cite{lang2021dpc}               & 20.82                               \\
SE-ORNET{[}U{]} ~\cite{deng2023se}          & 17.44                               \\
PointSetReg{[}A{]} ~\cite{zhao2024clustereg}       & 12.37                               \\
\rowcolor[HTML]{E7E6E6} 
Ours{[}U{]}             & \textbf{6.19}                       \\ \hline
\end{tabular}
\end{table}

\subsection{Robustness}\label{sec:top}
Moreover, we evaluate the robustness of our model with respect to noise and rotation perturbation and report in Tab.~\ref{table:robust}. 
More specifically, we perturb the point clouds by: 1) Adding per-point Gaussian noise with i.i.d $\mathcal{N}(0, 0.02)$ along the normal direction on each point; 2) Randomly rotating $\pm 30$ degree along some randomly sampled direction. 
We perform $3$ rounds of test, and report both mean error and the standard deviation in parentheses. 
Our pipeline delivers the most robust performance among all the other baselines (\textbf{0.1} vs. 0.11, \textbf{0.25} vs. 0.41), including SE-ORNET\cite{deng2023se} which is designed for rotational robustness. 
\begin{table}[t!]
    \caption{Mean geodesic errors $(\times100)$ on under different perturbations. Noisy PC means the input point clouds are perturbed by Gaussian noise. Rotated PC means the input point clouds are randomly rotated within ±30 degrees. The standard deviation value is shown in parentheses.
    }\label{table:robust}
    \centering
    \scriptsize
    \begin{tabular}{lcccc}
    \hline
    \textbf{Method}   & \textbf{}                                    & \textbf{Unperturbed} & \textbf{Noisy PC} & \textbf{Rotated PC} \\ \hline
    DiffFMaps{[}S{]}~\cite{marin2020correspondence} &                                              & 12.0                 & 14.9(2.57)        & 26.5(3.35)          \\
    NIE{[}U{]}~\cite{nie}       &                                              & 11.0                 & 11.5(0.32)        & 19.9(1.29)          \\
    SSMSM{[}U{]}~\cite{cao2023}     & \multirow{-3}{*}{\begin{tabular}[c]{@{}c@{}}Mesh\\ Required\end{tabular}}                        & 4.1                  & 5.4(0.11)         & 9.2(1.01)           \\ \hline
    DPC{[}U{]}~\cite{lang2021dpc}       &                                              & 17.3                 & 18.2(0.80)        & 22.1(0.72)          \\
    SE-ORNet{[}U{]}~\cite{deng2023se}  & \multirow{-2}{*}{\begin{tabular}[c]{@{}c@{}}Pure\\ PCD\end{tabular}}                 & 24.6                 & 24.7(0.15)        & 27.2(0.41)          \\
    \rowcolor[HTML]{E7E6E6} 
    Ours {[}U{]}      & \multicolumn{1}{l}{\cellcolor[HTML]{E7E6E6}} & \textbf{6.2}                 & \textbf{6.4}(\textbf{0.10})         & \textbf{7.0}(\textbf{0.25})           \\ \hline
    \end{tabular}\vspace{-1em}
\end{table}

\subsection{More Visualizations}\label{sec:performance}
\noindent\textbf{High-dimensional feature visualization:} 
To further validate the characteristics of the representations learned by our method, we present a set of more comprehensive visualizations of the features. 
As shown in Fig.~\ref{fig:feature}, our feature distribution is more clean and localized.
However, upon losing geometric or semantic information, the features across different dimensions become divergent, resulting in the loss of regular fine-grained representation at various levels.

\begin{figure}[!t]
    \centering
    \includegraphics[width=0.4\textwidth]{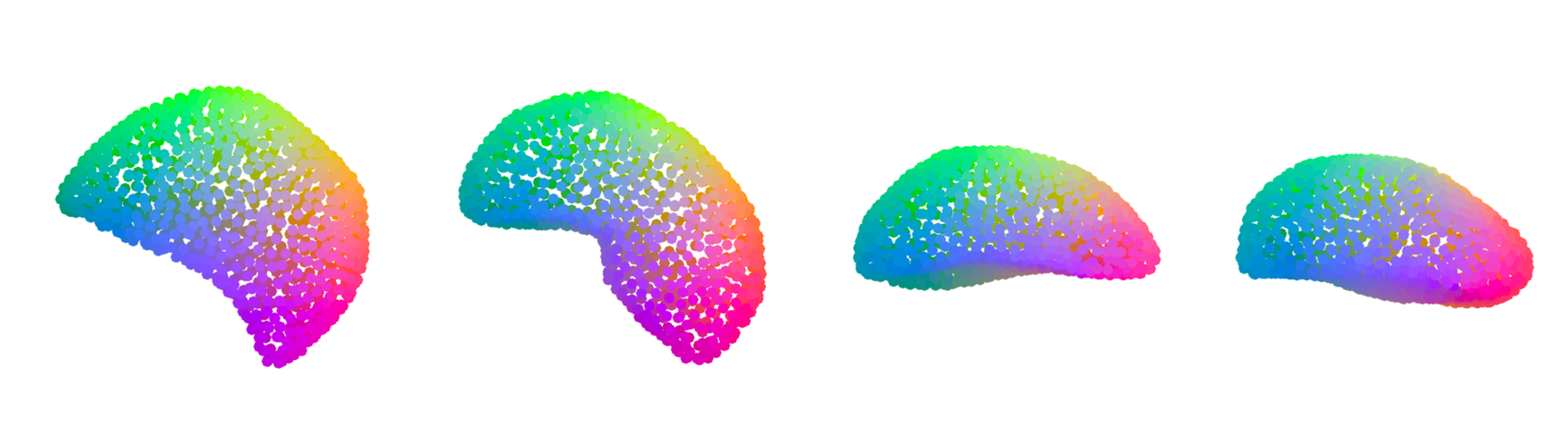}
    \vspace{-1em}
    \caption{Our matching result of the spleen dataset from~\cite{adams2023point2ssm}. 
    }\label{fig:spleen}
\end{figure}

\noindent\textbf{Matching results of medical dataset:} 
To supplement Tab.~\ref{table:med} of the main text, we further visualize the matching results on the Spleen in Fig.\ref{fig:spleen}, where excellent mapping is achieved regardless of whether the spleen exhibits various shapes or is positioned at different angles.

\begin{figure}[!t]
    \centering
    \includegraphics[width=0.5\textwidth]{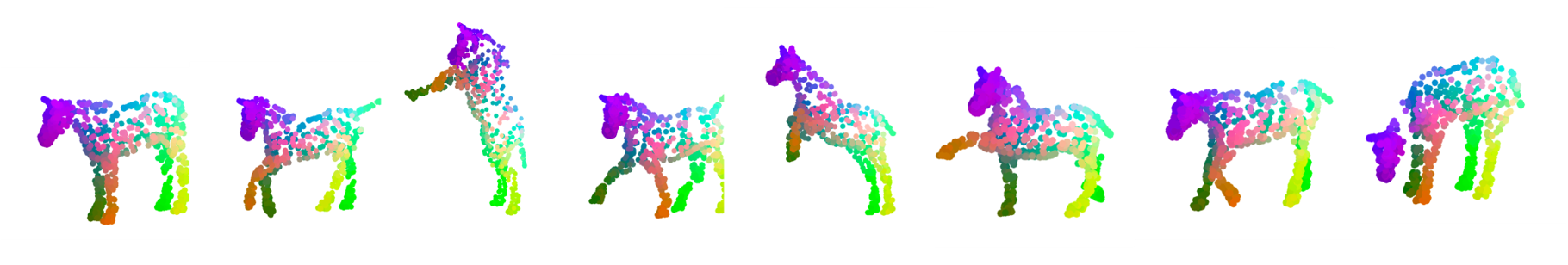}
    \vspace{-1.5em}
    \caption{More qualitative results of TOSCA. All horse shapes from the dataset have been showcased.
    }\label{fig:horse}
    \vspace{-0em}
\end{figure}

\begin{figure}[!t]
    \centering
    \includegraphics[width=0.5\textwidth]{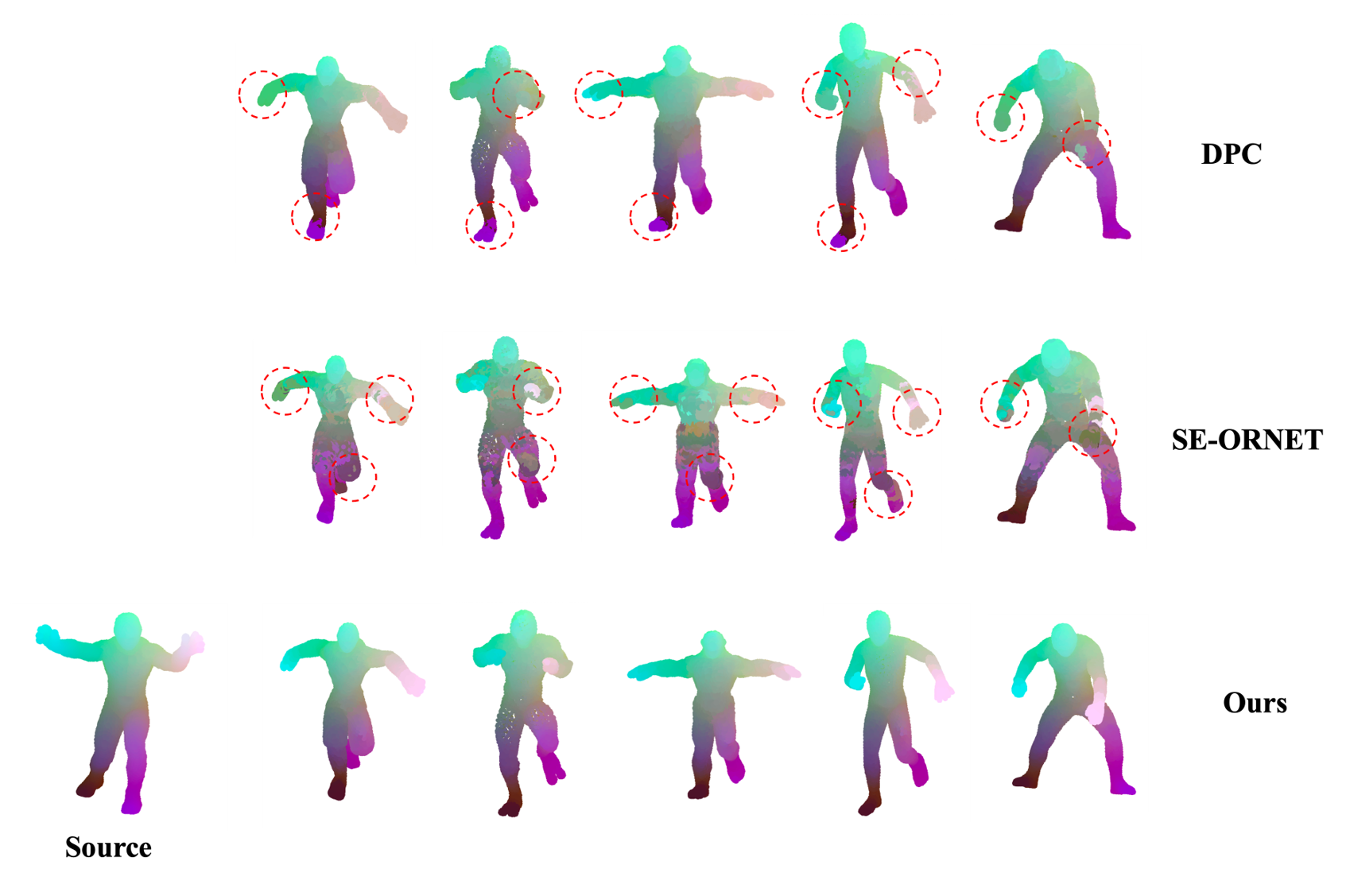}
    \vspace{-1.5em}
    \caption{More qualitative results of DT4D. Our method demonstrates a notable improvement over other baselines.
    }\label{fig:dt4d}
    \vspace{-1em}
\end{figure}

\begin{figure}[!t]
    \centering
    \includegraphics[width=0.5\textwidth]{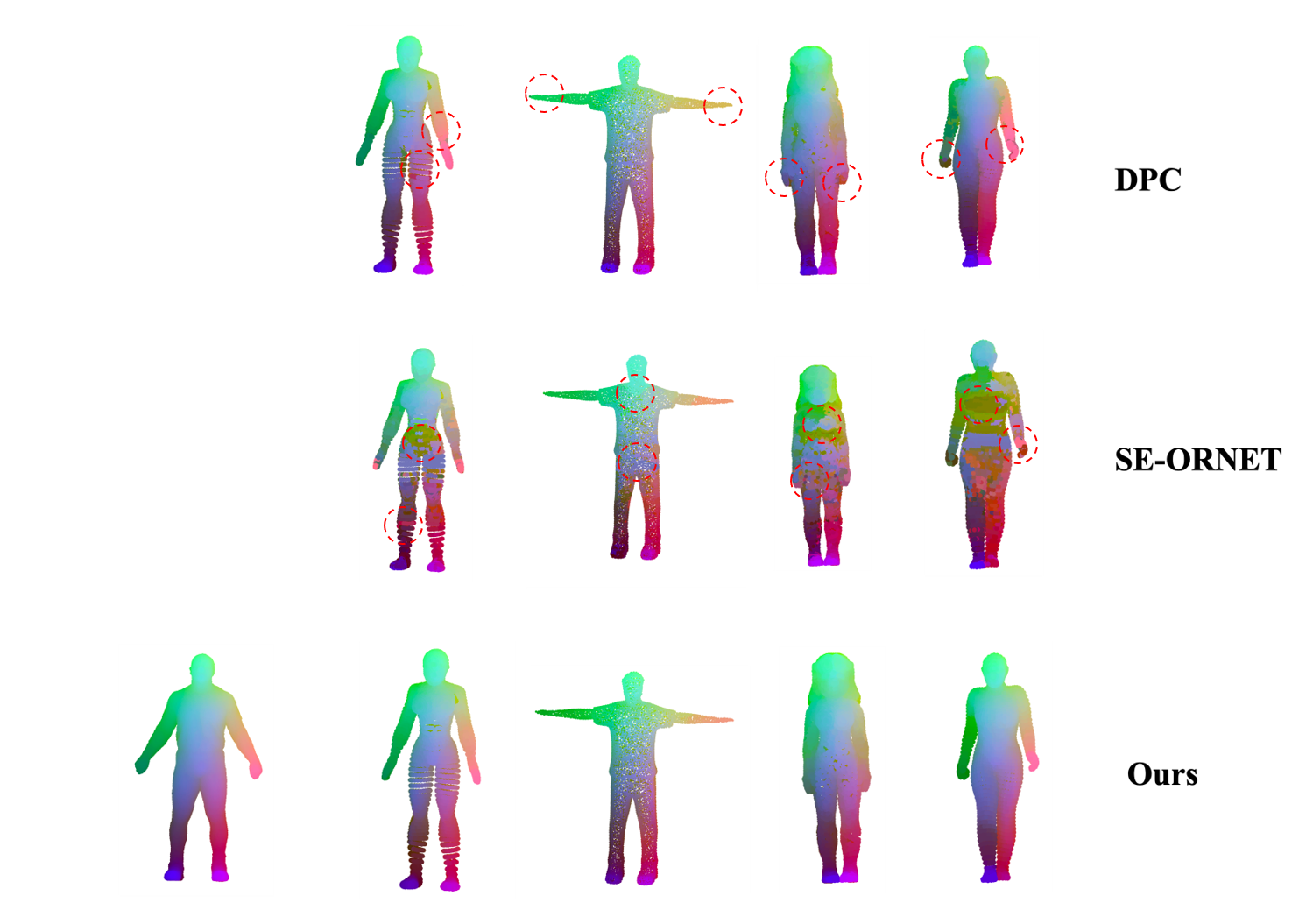}
    \vspace{-1.5em}
    \caption{More qualitative results of SHREC'07. Our approach significantly outperforms other baselines.
    }\label{fig:shrec07}
    \vspace{-0em}
\end{figure}

\begin{figure}[!t]
    \centering
    \includegraphics[width=0.5\textwidth]{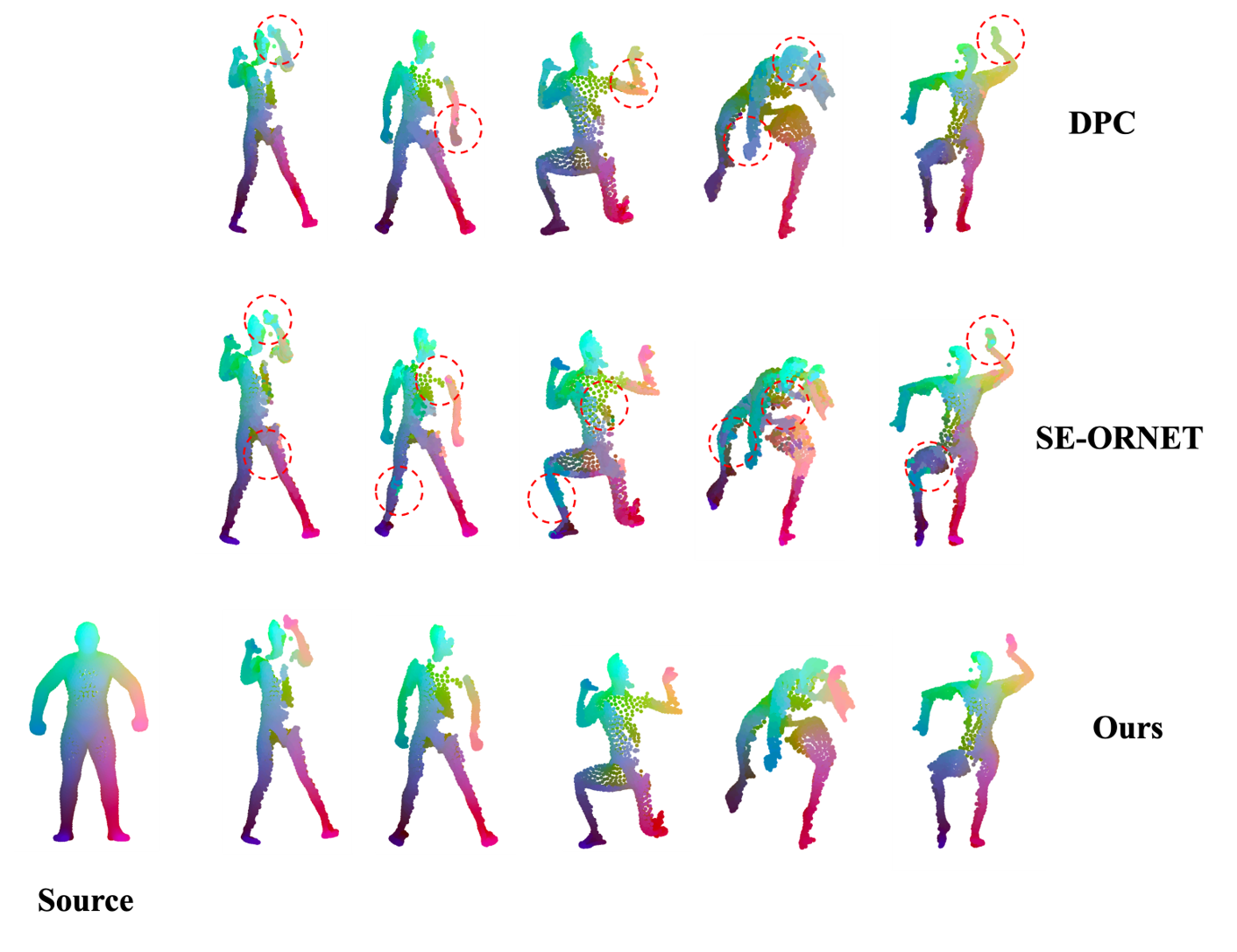}
    \vspace{-1.5em}
    \caption{More qualitative results of SCPAE-PV. Our approach achieves superior performance over other baselines across various partial views.
    }\label{fig:scape_pv}
    \vspace{-1em}
\end{figure}

\begin{figure}[!t]
    \centering
    \includegraphics[width=0.5\textwidth]{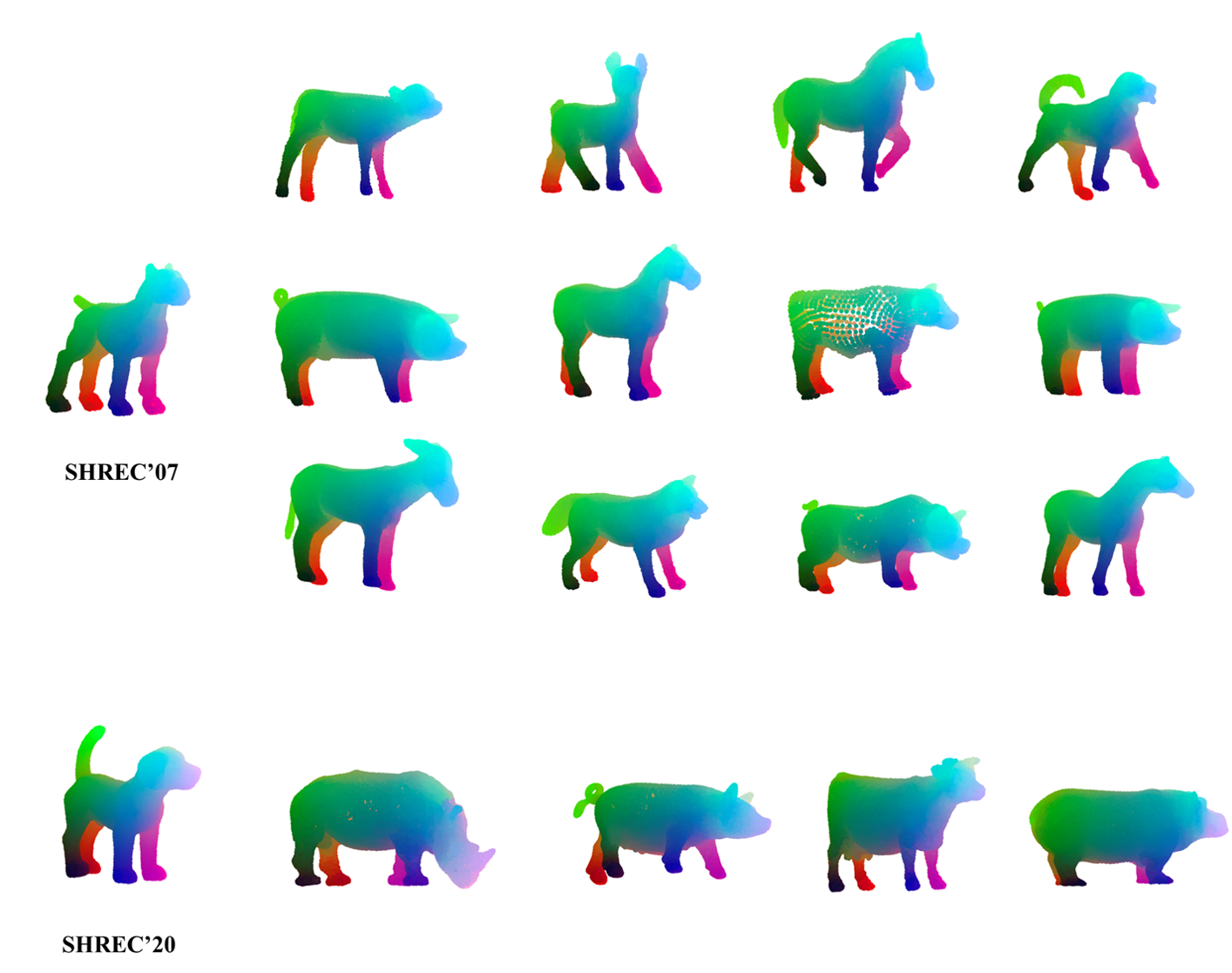}
    \vspace{-1.5em}
    \caption{More qualitative results of SHREC'07-Fourleg and SHREC'20.
    }\label{fig:shrecours}
    \vspace{-1em}
\end{figure}

\noindent\textbf{More qualitative results:} 
We further visualize the results of TOSCA, DT4D, SHREC'07 and SCAPE-PV, which respectively serve as qualitative validation supplements for learning sparse point clouds in Tab.~\ref{table:baseline}, the generalization capability in Tab.~\ref{table:iso_noiso} of the main text, and the adaptability to partial shapes in Tab.~\ref{table:partial} of the main text. 
The training and testing procedures align with the methods described in the aforementioned table, with quantitative supplements presented respectively in Fig.~\ref{fig:horse}, Fig.~\ref{fig:dt4d}, Fig.~\ref{fig:shrec07} and Fig.~\ref{fig:scape_pv}, respectively. 
Furthermore, to supplement Fig.~\ref{fig:shrec} and Tab.~\ref{table:fourleg}, we further visualized the quantitative performance of our method on SHREC'07-Fourleg and SHREC'20 in Fig.~\ref{fig:shrecours}.
\begin{table*}[!t]
    \caption{Hyper-parameters. The tables details the hyperparameter values that we used for the training of SCAPE\_r.}\label{table:hyber}
    \centering
    \scriptsize
    \begin{tabular}{llc}
    \hline
    \textbf{Symbol} & \textbf{Description}                                                                                                                         & \textbf{Value} \\ \hline
    k\_dist         & The nearest number for computing geometrically similarity loss.                                                                              & 500            \\
    N\_dist         & The number of points sampled to calculate the geometrically similarity loss.                                                                 & 1000           \\
    k\_deform       & The number of neighborhood features gathered in the graph convolutional network of our Deformer.                                             & 10             \\
    k\_attn         & The number for searching latent nearest features in our Local Attention Block.                                                               & 40             \\
    C               & The dimention of output feature.                                                                                                             & 128            \\
    decay\_factor   & The multiplicative factor by which the learning rate is reduced during each decay.                                                           & 0.5            \\
    decay\_iter     & The epoch interval at which the learning rate is decayed.                                                                                    & 10             \\
    alpha           & The temperature parameter in the Softmax function. It is a dynamically increasing value with epochs. As alpha increases,  Pi becomes harder. & 10-100         \\
    TEs             & Training epochs.                                                                                                                             & 20             \\
    H,W             & The size of our projected image.                                                                                                             & 224,224        \\ \hline
    \end{tabular}
\end{table*}
\subsection{Experimental Setup}\label{sec:env}
We perform all the experiments on a machine with NVIDIA A100-SMX4 80GB and Intel(R) Xeon(R) CPU E5-2680 v4 @ 2.40GHz using the PyTorch 2.2.0 framework. 

\subsection{Additional Hyper-parameter Details}\label{sec:hyper}
For a comprehensive understanding of the specific hyper-parameter configurations, please refer to Tab. \ref{table:hyber}.

\section{Broader Impacts}\label{sec:bro}
We fail to see any immediate ethical issue with the proposed method. On the other hand, since our method is extensively evaluated in matching human shapes and achieves excellent results, one potential misuse can be surveillance, which may pose negative societal impact. 

\end{document}